%% file: paper.tex
\documentclass[10pt,letterpaper]{paper}

\usepackage[all]{hypcap}

\usepackage[authoryear, round]{natbib}

\usepackage[utf8]{inputenc} 
\usepackage[T1]{fontenc}    
\usepackage{url}            
\usepackage{booktabs}       
\usepackage{amsfonts}       
\usepackage{nicefrac}       
\usepackage{microtype}      
\usepackage{multirow} 
\usepackage[textsize=tiny]{todonotes}
\usepackage{algorithm}
\usepackage{amssymb}
\usepackage{cleveref}
\usepackage{dsfont}
\usepackage{nicefrac}
\usepackage{inconsolata}
\usepackage{xcolor}
\usepackage{graphicx}
\usepackage{amsmath}
\usepackage{amssymb}
\usepackage{booktabs}
\usepackage{multirow}
\usepackage{makecell}
\usepackage{caption}
\usepackage{subcaption}
\usepackage{bbm}
\usepackage{wrapfig}
\usepackage{mathtools,xparse}
\usepackage{pifont}
\usepackage{enumitem}
\usepackage{array}
\usepackage{scalerel,xparse}
\usepackage{colortbl}
\usepackage{tablefootnote}
\usepackage{tocloft}  
\usepackage{hyperref}  
\usepackage{cleveref}
\usepackage{dblfloatfix} 
\usepackage{adjustbox}
\usepackage{algpseudocode}
\usepackage{setspace}
\usepackage[most,skins,theorems]{tcolorbox}
\usepackage[title,toc,titletoc]{appendix}
\usepackage{natbib}
\usepackage{thm-restate}
\usepackage{amssymb}
\usepackage{amsfonts}
\usepackage{ulem}
\usepackage{tocloft}
\newboolean{showsection}
\setboolean{showsection}{true}
\makeatletter
\@namedef{ver@everyshi.sty}{}
\makeatother

\definecolor{custom_green}{rgb}{0.0, 0.5, 0.0}
\definecolor{custom_red}{rgb}{1.0, 0.01, 0.24}
\definecolor{custom_blue}{HTML}{C9DAF7}
\definecolor{custom_purple}{HTML}{D9D1E9}
\definecolor{title_blue}{HTML}{204899} 
\definecolor{cite_blue}{HTML}{044dc1}  
\definecolor{cite_purple}{HTML}{7406a7}  
\newcommand*{\pd}[3][]{\ensuremath{\frac{\partial^{#1} #2}{\partial #3}}}

\hypersetup{
    colorlinks = true,
    citecolor = {cite_blue},
    linkcolor = {cite_purple},
    urlcolor = {cite_purple},
}

\input{macro}

\input{math_commands.tex}

\tcbset{
    positive/.style={
        colback=custom_blue!5!white, 
        colframe=custom_blue!60!black, 
        coltitle=custom_blue!50!black, 
        colbacktitle=custom_blue!20!white, 
        fonttitle=\bfseries\centering, 
        title=Positive Steering,
        rounded corners,
        boxrule=0.5mm,
        enhanced,
        before skip=5mm, 
        after skip=5mm,  
    },
    neutral/.style={
        colback=gray!5!white, 
        colframe=black!50!white, 
        coltitle=black!50, 
        colbacktitle=black!10!white, 
        fonttitle=\bfseries\centering, 
        title=Neutral Response,
        rounded corners,
        boxrule=0.5mm,
        enhanced,
        before skip=5mm,
        after skip=5mm,
    },
    negative/.style={
        colback=custom_orange!5!white, 
        colframe=custom_orange!60!black, 
        coltitle=custom_orange!50!black, 
        colbacktitle=custom_orange!20!white, 
        fonttitle=\bfseries\centering, 
        title=Negative Steering,
        rounded corners,
        boxrule=0.5mm,
        enhanced,
        before skip=5mm,
        after skip=5mm,
    },
    dataexamplebox/.style={
        colframe=ccbgc_2,
        colback=ccbgc,
        coltitle=black,
        fonttitle=\small\bfseries,
        colbacktitle=ccbgc_2,
        title=Example: Myopic Alice/Bob,
        rounded corners,
        boxrule=0.5mm,
        arc=1mm,
        width=\columnwidth,
        top=1mm,
        bottom=1mm,
        left=1mm,
        right=1mm,
        fontupper=\footnotesize
    }
}

\newcommand\pythonstyle{\lstset{
basicstyle=\ttfamily\footnotesize,
language=Python,
morekeywords={self, clip, exp, mse_loss, uniform_sample, concatenate, logsumexp},              %
keywordstyle=\color{deepblue},
emph={MyClass,__init__},          %
emphstyle=\color{deepred},    %
stringstyle=\color{deepgreen},
frame=single,                         %
showstringspaces=false
}}

\lstnewenvironment{python}[1][]
{
\pythonstyle
\lstset{#1}
}
{}

\newcommand\pythoninline[1]{{\pythonstyle\lstinline!#1!}}

\makeatletter
\def\mathcolor#1#{\@mathcolor{#1}}
\def\@mathcolor#1#2#3{%
  \protect\leavevmode
  \begingroup
    \color#1{#2}#3%
  \endgroup
}
\makeatother

\tcbset{
  aibox/.style={
    width=474.18663pt,
    top=7pt,
    bottom=5pt,
    colback=blue!6!white,
    colframe=black,
    colbacktitle=black,
    enhanced,
    center,
    attach boxed title to top left={yshift=-0.1in,xshift=0.15in},
    boxed title style={boxrule=0pt,colframe=white,},
  }
}
\newtcolorbox{AIbox}[2][]{aibox,title=#2,#1}

\Crefformat{equation}{#2Eq.\;(#1)#3}

\Crefformat{figure}{#2Figure #1#3}
\Crefformat{assumption}{#2Assumption #1#3}
\Crefname{assumption}{Assumption}{Assumptions}

\usepackage{crossreftools}
\newtheorem*{lemma*}{Lemma}

\makeatletter
\renewcommand\footnoterule{%
  \kern 15\p@
  \hrule \@width 2in \kern 2.6\p@ 
  \vspace{4pt}
}
\makeatother

\usepackage{titlesec}
\titleclass{\subsubsubsection}{straight}[\subsubsection]
\newcounter{subsubsubsection}[subsubsection]
\renewcommand\thesubsubsubsection{\thesubsubsection.\arabic{subsubsubsection}}
\titleformat{\subsubsubsection}
  {\normalfont\normalsize\bfseries}{\thesubsubsubsection}{1em}{}
\titlespacing*{\subsubsubsection}{0pt}{3.25ex plus 1ex minus .2ex}{1.5ex plus .2ex}

\title{How Smoothing is N-simplicial Attention ?}
\reportnumber{} %

\author[1,2]{Alexandre Dussolle}
\author[1]{Pietro Li\`o}

\affil[1]{University of Cambridge}
\affil[2]{École des Ponts, IP Paris}

\correspondingauthor{Correspondence to: \email{agld2@cam.ac.uk} (Alexandre Dussolle)}

\begin{abstract}
\textbf{Abstract:} 
Going from pure Multilayer Perceptron (MLP) to a learnable graph message-passing mechanism at each layer has been foundational to state-of-the-art results, despite the computational trade-off (e.g. GATs or Transformers). To go a step further, in this work, we introduce N-simplicial attention, going from pairwise token similarity to higher-order interactions, and adapt it for Rotary Position Embeddings (RoPE). To help manage the increased complexity, we propose a cost-effective simplex selection enabling the model to focus its computation load onto the more task-sensitive interactions. Beyond these core mechanisms, we study how smoothing N-simplicial attention is by deriving a Lipschitz upper-bound and by demonstrating that by itself it also suffers from over-smoothing, despite opening the attention message-passing to higher-order interactions.
\end{abstract}

\begin{document}

\maketitle
\input{tex/part1}

\input{tex/part3}
\input{tex/part4}
\input{tex/part2}

\clearpage
\bibliographystyle{plainnat}
\bibliography{paper.bib}

\newpage

\setlength{\cftbeforesecskip}{16pt}
\setlength{\cftbeforesubsecskip}{4pt}
\renewcommand\cftsecpagefont{\color{RoyalBlue}}
\renewcommand\cftsubsecpagefont{\color{RoyalBlue}}
\renewcommand\cftsubsubsecpagefont{\color{RoyalBlue}}
\renewcommand{\contentsname}{\large{Contents}}
{
  \hypersetup{linkcolor=}
  \tableofcontents
}
\renewcommand{\appendixname}{}

\begin{appendices}
    \input{tex/annex2}
    \input{tex/annex3}

    \input{tex/annex5}

    \input{tex/annex4}
\end{appendices}

\end{document}

%% file: macro.tex
\definecolor{blanchedalmond}{rgb}{1.0, 0.92, 0.8}
\definecolor{carmine}{rgb}{0.59, 0.0, 0.09}
\definecolor{lightblue}{rgb}{0.22,0.45,0.70}%

\newtheorem{theorem}{Theorem}[section]

\newtheorem{lemma}[theorem]{Lemma}
\crefname{lemma}{Lemma}{Lemmas}
\newtheorem{corollary}[theorem]{Corollary}

\newtheorem{definition}[theorem]{Definition}

\newtheorem{remark}[theorem]{Remark}

\crefformat{lemma}{#2Lemma~#1#3}
\Crefformat{lemma}{#2Lemma~#1#3}

\newtheorem*{definition*}{Definition}
\newtheorem*{remarks*}{Remarks}

\renewcommand{\mathbf}{\boldsymbol}

\makeatletter
\def\Ddots{\mathinner{\mkern1mu\raise\p@
\vbox{\kern7\p@\hbox{.}}\mkern2mu
\raise4\p@\hbox{.}\mkern2mu\raise7\p@\hbox{.}\mkern1mu}}
\makeatother

\numberwithin{equation}{section}

\definecolor{amaranth}{rgb}{0.9, 0.17, 0.31}
\definecolor{antiquebrass}{rgb}{0.8, 0.58, 0.46}
\definecolor{antiquefuchsia}{rgb}{0.57, 0.36, 0.51}
\definecolor{chromeyellow}{rgb}{0.31, 0.47, 0.26}

\newcommand{\1}{\mathds 1}


%% file: math_commands.tex
\usepackage{amsmath,amsfonts,bm}

\def\eqref#1{equation~\ref{#1}}

\def\1{\bm{1}}

\DeclareMathAlphabet{\mathsfit}{\encodingdefault}{\sfdefault}{m}{sl}
\SetMathAlphabet{\mathsfit}{bold}{\encodingdefault}{\sfdefault}{bx}{n}



%% file: tex/part1.tex
\section{Introduction} \label{sec:intro}
Attention is a central part of the transformer architecture \citep{vaswani2023attentionneed}, which has become fundamental to modern deep learning models, for language (\citealt{touvron2023llamaopenefficientfoundation}, 
\citealt{openai2024gpt4technicalreport}, \citealt{gemmateam2024gemmaopenmodelsbased}, \citealt{deepseekai2025deepseekv3technicalreport}), for vision (\citealt{dosovitskiy2021imageworth16x16words}, \citealt{radford2021learningtransferablevisualmodels}, \citealt{oquab2024dinov2learningrobustvisual}), and partly for graphs as well (\citealt{ying2021transformersreallyperformbad}, \citealt{joshi2025transformersgraphneuralnetworks}, \citealt{Buterez_2025}, \citealt{bechlerspeicher2025positiongraphlearninglose}). 
Unlike MLPs, which treat each token independently, attention allows the model to learn more explicitly how their interactions should influence its output, by considering pairwise connections at each layer and letting the depth construct more complicated interactions. This is part of a broader graph message-passing framework (GNN: \citealt{10.5555/2987189.2987330}, \citealt{Gori2005ANM}, \citealt{4700287}, GAT: \citealt{veličković2018graphattentionnetworks}, and attention as graph message-passing \citealt{joshi2025transformersgraphneuralnetworks}). To adapt to specific data natural topological structure, previous works developed higher-order models (\citealt{morris2021weisfeilerlemanneuralhigherorder}, \citealt{goh2022simplicialattentionnetworks}, \citealt{hajij2023topologicaldeeplearninggoing}) working directly on the given presupposed structure and keeping higher-order features in memory (see \autoref{appx:hom} for details). In parallel, the molecules geometry led to slight (sometimes cubic) variations of attention 
 (\citealt{Jumper2021}, \citealt{liu2022gem2generationmolecularproperty}, \citealt{hussain2024tripletinteractionimprovesgraph}).

\noindent In this work, we introduce \textit{N-simplicial attention}, generalizing attention to any higher-order interactions between tokens. This expands from previous works on 2-simplicial attention (\citealt{clift2019logic2simplicialtransformer},  \citealt{roy2025fastsimplex2simplicialattention}), and on cubic attention (\citealt{sanford2023representationalstrengthslimitationstransformers}, \citealt{alman2023capturehigherordercorrelationsgeneralizing}, \citealt{liang2024tensorattentiontrainingprovably}), that describe the emerging trade-off between increasing to cubic complexity and the improvements we get on the scaling laws exponents and on the needed precision scaling. Which could be beneficial as the limited total available data imply a growing need for architectures that are more token-input efficient and capable of focusing their compute load better. 
One way to do this adaptive computation is from early-exiting simpler tokens (\citealt{elbayad2020depthadaptivetransformer}, \citealt{schuster2022confidentadaptivelanguagemodeling}, \citealt{Elhoushi_2024}), we also introduce a more fine-grained selection system allowing for sparse higher-order attention. Furthermore, we highlight how \textit{N-simplicial attention} is (like attention) also prone to the over-smoothing phenomena, which put previous work on complexity reduction (\citealt{liang2024tensorattentiontrainingprovably}) into perspective as they required bounded entries which accentuates the model over-smoothing tendencies.

\noindent \textbf{Contributions. } In brief, our main contributions in this paper are as follows,
\vspace{-9pt}
\begin{itemize}[leftmargin=*, itemsep=2pt]
    \item \textbf{N-simplicial attention:}  we present a generalization of attention to higher-order token   interactions \S \ref{subsec:nsimp} 

     \item \textbf{Sparse simplicial attention:} we introduce a router selecting which token simplex is to be considered by the model, to reduce the complexity, the router is light-weight, trainable and inference-time compatible. \S \ref{subsec:router}

     \item \textbf{Simplicial Rotary Embeddings:} we adapt N-simplicial attention to support RoPE (\citealt{su2023roformerenhancedtransformerrotary}) \S \ref{subsec:rope}
    
    \item \textbf{Over-smoothing affects N-simplicial attention:} we demonstrate that naively stacking N-simplicial attention layers leads to the rank-collapse and over-smoothing of the tokens representation (with or without mask) in a similar way attention and GNN message-passing do. 
    \S\ref{subsec:ovsmN}

    \item \textbf{Lipschitz upper bound for N-simplicial attention:} we show how going from attention to higher-order attention affects its \textit{local} Lipschitz continuity, by studying an upper bound of the Lipschitz constant, translating how smoothing the layer will be for two different but relatively close inputs. \S\ref{subsec:upper}
    
\end{itemize}

\newpage
\section{Methods}
\subsection{Preliminary}
\noindent
Usually, the attention logits are $QK^T =  \big( Q \otimes K \big)_{ia\,ja} \quad$ with $Q = XW_Q \quad K=XW_K $ \\ with here $X : \{ n\times d\}$ the tokens representation (batch and heads are omitted below for readability)\\
\noindent
recalled from \cite{vaswani2023attentionneed}, also writable with a tensor product and n-mode product :
\begin{equation*}
    QK^T = \, \Big( \overset{d}{\underset{a=1}{\sum}} Q_{i a}\, K_{j a} \Big)_{i,j\, \leq \,n} \,=  \big( Q \otimes K \big)_{ia\,ja} = \big( Q \otimes K \big) \times_{2} \underline{1}_{\,d}^{T} \times_{4} \underline{1}_{\,d}^{T}
\end{equation*}
$Q \otimes K$ is a $n\times d\times n \times d$ tensor, then contracted into the $n\times n$ attention logits matrix. \hfill (see \autoref{fig:attn})$\quad \; \;$\\
Or in the following pseudocode, with Einstein notation :
logits $\gets$ \texttt{einsum("bhqd, bhkd $\to$ bhqk", Q, K)} \\
For clarity, the logits' Einstein notation without batch and head : $qd, kd \to qk$, and the whole algorithm:  

\begin{algorithm}
\caption{Pseudocode for the forward pass of attention}
\begin{algorithmic}[1]
\State \textbf{Procedure} ATTENTION(Q, K, V)
\State logits $\gets$ \texttt{einsum("bhqd, bhkd $\to$ bhqk", Q, K)}
\State attention $\gets$ \texttt{softmax(logits + mask, axis = [-1])}
\State output $\gets$ \texttt{einsum("bhqk, bhkd $\to$ bhqd", attention, V)}
\State \textbf{Return} output
\end{algorithmic}
\end{algorithm}
\subsection{N-simplicial attention} \label{subsec:nsimp} \label{ref:algo}
\noindent Attention only considers pairwise interactions between tokens at each step, to go further, taking inspiration on 2-simplicial attention (\citealt{clift2019logic2simplicialtransformer}, \citealt{roy2025fastsimplex2simplicialattention}),
it generalizes to the N-simplicial case \citep{pappone2025beyondattention} a 0-simplexe is a node, 1-simplexe is an edge, 2-simplexe a triangle, etc. (see \autoref{fig:placeholder})\\
\noindent
The $\{n\times n\}$ attention logits' matrix becomes a $n^{N+1}$ tensor : 
$\big( Q \otimes K \big)_{ia\,ja} \underset{}{\to} (Q \otimes K_1 \otimes \cdots \otimes K_N)_{qa\,k_1a\,\cdots\, k_Na}$
\begin{equation*}
\Big(K_0 \otimes \cdots \otimes K_N \Big)_{k_0a \,k_1 a \,\cdots \, k_N a}=
\left( \overset{N}{\underset{i=0}{\bigotimes}} \, K_i \right) \prod_{j=1}^{N} \times_{2j} \, \underline{1}_{\,d}^{T} \,\,\,\,\,= \,\, \Bigg(\, \overset{d}{\underset{a=1}{\sum}} (K_0)_{m_0a}\,\cdots\, (K_n)_{m_Na} \Bigg)_{1\leq m_{0},\ldots,m_{N}\leq n}
\end{equation*}
\noindent
$\underline{1}_{\,d}^{T} \in \mathbb{R}^{1\times d}$ : row vector of ones, and $\times_{k}$ : n-mode product along axis k (used for contraction) \\
\noindent
then use the softmax operator and project the attention score of each N-simplex back on $X$ using $V_1, \ldots, V_n$

\begin{equation*}
\mathcal{A} = \mathrm{softmax} \Bigg( \frac{1}{\sqrt{d}} \bigg( \overset{N}{\underset{i=0}{\bigotimes}} \, X W_{K}^{(i)} \bigg) \prod_{j=1}^{N} \times_{2j} \, \underline{1}_{\,d}^{T} \,\,\, , \,\mathrm{axis}=(1, \ldots,N) \Bigg)
\end{equation*}

\begin{equation*}
X'_{ij} = X_{ij} + \sum_{k_1, \ldots,k_N}  \mathcal{A}_{\,i\,k_1\cdots k_N} \, \cdot \, \prod_{m=1}^{N} \Big( X W_{V}^{(m)} \Big)_{ k_{m}j}
\end{equation*}
\noindent
Which give us the following pseudocode: \hfill (see \autoref{fig:2simp})
\begin{algorithm} 
\caption{Pseudocode for the forward pass of N-simplicial attention (without skip)}
\begin{algorithmic}[1]
\State \textbf{Procedure} N-SIMPLICIAL ATTENTION($K_0, \ldots,K_N$, $V_1,\ldots,V_N$)
\State logits $\gets$ \texttt{einsum("bhqd, bh$\text{k}_1$d,..., bh$\text{k}_{\text{N}}$d $\to$ bhq$\text{k}_1\cdots\text{k}_{\text{N}}$", $K_0, \ldots,K_N$)}
\State attention $\gets$ \texttt{softmax(logits, axis = [-1, ..., -N])}
\State output $\gets$ \texttt{einsum("bhq$\text{k}_1\cdots\text{k}_{\text{N}}$, bh$\text{k}_1$d,...,bh$\text{k}_{\text{N}}$d $\to$ bhqd", attention, $V_1,\ldots,V_N$)}
\State \textbf{Return} output
\end{algorithmic}
\end{algorithm} \\
For clarity, here's the logits' Einstein notation : $qd, k_1d, \ldots, k_nd \to qk_1\cdots k_n$ \hfill (without batch and multi-head) \\
\noindent
 Others non-linear multidimensional operators can be used instead of the multi-axis softmax (flatten all dimensions but the first), no issue with order of $XW^{(m)}_V$ as it's permutation invariant. \\
This also opens the door to "cross-attention" between more than 2 modalities.


\subsection{Simplicial token selection routing} \label{subsec:router}
 To lower the complexity, we preselect the tokens we apply N-simplicial attention on. Here's $2$ ways to do so:
 \\
 $\bullet$ \textbf{Expert-choice Top$\,$K tokens.}
 inspired by \citealt{bae2025mixtureofrecursionslearningdynamicrecursive}, to simulate learnable tokens early-exit behavior. \\
 we compute the score $(X_{i,:}^{(t)}\omega_{\theta} )$ for each token $X_{i,:}\in \mathbb{R}^{1\times d}$ using a learnable linear projection $\omega_{\theta}\in \mathbb{R}^{d\times 1}$ \\
 And we do the N-simplicial attention computation $f_t$ only for the top-K tokens, to which we add the results weighted by the score (to make it learnable). 
 $$
 X_{i,:}^{(t+1)} =
\begin{cases}
\displaystyle  X_{i,:}^{(t)} + s_i^{(t)}\, f_t(X_{i,:}^{(t)}) \quad& \text{if } \,\, s_i^{(t)} = X_{i,:}^{(t)}\omega_{\theta}  \geq \underset{j\leq n}{\text{Top$\,$K}} (s_j^{(t)} ) \\
\quad \quad X_{i,:}^{(t)} & \text{else.}
\end{cases}
$$
\noindent 
Although this guarantees perfect load balancing (\citealt{raposo2024mixtureofdepthsdynamicallyallocatingcompute}, \citealt{bae2025mixtureofrecursionslearningdynamicrecursive}), in the sequential case it suffers from information leakage (\citealt{zhou2022mixtureofexpertsexpertchoicerouting}, \citealt{wang2024auxiliarylossfreeloadbalancingstrategy}) as it's comparing all tokens without regard for causality. 
In the graph case however, there is no causality constraint for the tokens (nodes), therefore no leakage issue: we have the benefits of the expert-choice routing without its major weakness. \\

\noindent 
$\bullet$ \textbf{Simplicial Path Sparse attention:} while expert-choice preselects $m\leq n$ tokens, to do N-simplicial attention only on those, we now want to look into a method that would be more "fine-grained" and that doesn't break the causality mask constraint. In Deepseek Sparse Attention (DSA) \citep{deepseekai2024deepseekv32} after computing a score for each pair of tokens using the previous attention logits, we select for each token the corresponding top-K others, through the score (for each head, then sum over them): $ m_{ij} = \mathbf{1}_{\{ s_{i,j} \notin\,  {\text{Top$\,$K}}_{(a\leq N)}(s_{i,a}) \} } $ $$
s_{i,j}^{(t)} = w_{i} \,\text{ReLU}(\langle Q_{i,:}^{(t-1)} , K_{j,:}^{(t-1)} \rangle)
\qquad \qquad \Big((i,j) \in E(\mathcal{G}),\,\, \mathcal{G} \text{ : causal mask}\Big)$$
for each head, with $Q,K$ (query, key) obtained from the previous attention layer, $w_i:$ linear projection of $X_{i,:}$. \\
Adapting this naively to N-simplicial attention would mean a score for each N-simplex, compute expensive. 
$ s_{m_0,\ldots,m_N}^{(t)} = w_{m_0} \,\text{ReLU}(\langle K_{0_{m_0,:}}^{(t-1)}, \ldots, K_{N_{m_N,:}}^{(t-1)} \rangle) \quad $ with $\langle x^{(0)}, \ldots, x^{(N)} \rangle = \underset{a=1}{\overset{d}{\sum}} x^{(0)}_a\,\cdots\, x^{(N)}_a$ \\
To avoid this excessive cost, we want to choose the simplexes by using only the DSA pairwise scores, \\
we introduce \textit{Simplicial Path Sparse Attention} where the $nk^{N-1}$ top simplexes are selected by following all the possible paths of the top-k tokens at each node. The simplicial mask is:  $ M_{k_0 \cdots k_N} = \mathbf{1}_{\{ (\underset{j \leq n}{\bigvee} m_{k_j k_{j+1}}) \, = 1 \}}$ i.e. $$ \text{the simplex }(k_0, \cdots, k_N) \text{ is selected if }\,\,\, \forall i\leq N, s_{k_i,k_{i+1}} \in \underset{j\leq n}{\text{Top$\,$K}}(s_{k_i,j}) $$ 


\subsection{Rotary Positional Embeddings (RoPE) for N-simplicial attention} \label{subsec:rope}
We propose here to adapt RoPE \citep{su2023roformerenhancedtransformerrotary} for the N-simplicial case. Commonly used to capture positional information in a sequence (\citealt{touvron2023llamaopenefficientfoundation}, \citealt{gemmateam2024gemmaopenmodelsbased}, \citealt{deepseekai2025deepseekv3technicalreport}), RoPE applies position dependent rotations to the query $q_i$ and the key $k_j$ making the dot product $\langle q_i, k_j \rangle$ depends on the relative distance $i-j$. Note that it's invariant under the same orthogonal transformation $R\in O(d), \,\, \langle R q_i, R k_j \rangle = \langle q_i, k_j \rangle$, useful as we expect the dot product of query and key at the same position to be unchanged. \\
Sadly the simplicial attention N-linear form $\langle x^{(0)}\,, \ldots, x^{(N)} \rangle = \overset{d}{\underset{a=1}{\sum}} x^{(0)}_{a} \cdots x^{(N)}_{a}$ is not invariant under rotation. To generalize RoPE to N-simplicial attention (as \citealt{roy2025fastsimplex2simplicialattention} did for N=2) we define the new logits as : 
$$
\mathcal{L}_{m_0, \cdots,m_N}^{(det)} = \underset{a=1}{\overset{\lfloor\frac{d}{N}\rfloor}{\sum}} \det \Big( \left[ (K_0)_{m_0,:}^{(a,N+1)}, \cdots, (K_N)_{m_N,:}^{(a,N+1)} \right] \Big)
$$ 
with $x^{(a,N)} = x[N(a-1), Na]$, the $a^{th}$ chunk of size N of $x$. Now rotation invariant, we can use RoPE as usual. \\ 
To finish the logits are then softmaxed and projected on $X$ using the values matrices as we previously did. \S\ref{subsec:nsimp}
Because the study of positional encoding is still an active field (NoPE: \citealt{kazemnejad2023impactpositionalencodinglength}, \citealt{wang2024lengthgeneralizationcausaltransformers}) with unanticipated effects \citep{barbero2025roundroundgomakes}, we decided to present in this paper the two different version of N-simplicial attention above (\S\ref{subsec:nsimp} and \S\ref{subsec:rope}).

%% file: tex/part3.tex
\section{Theoretical analysis}
As N-simplicial attention shifts the message-passing topology (from graph to hypergraph) and alleviate depth directly into the attention mechanism, we propose below a theoretical analysis focused on what they impact : the over-smoothing, over-squashing phenomena and the Lipschitz constant. \\
\textbf{Over-smoothing} (for GNN: \citealt{rusch2023surveyoversmoothinggraphneural}, \citealt{keriven2022littlemuchtheoreticalanalysis}, for GAT: \citealt{wu2024demystifyingoversmoothingattentionbasedgraph}) also called rank-collapse (\citealt{roth2024rankcollapsecausesoversmoothing}, for decoder-only transformers: \citealt{noci2022signalpropagationtransformerstheoretical},  \cite{dong2023attentionneedpureattention}, \citealt{wu2024roleattentionmaskslayernorm}) describes the convergence of all the tokens to a single representation, losing potential information. \\
\textbf{Over-squashing} (\citealt{alon2021bottleneckgraphneuralnetworks},
\citealt{digiovanni2023oversquashingmessagepassingneural}, \citealt{barbero2025llmsattendtoken}) occurs when information isn't efficiently propagated throughout the graph, making a token representation insensitive to  distant nodes.

\subsection{Over-smoothing/over-squashing trade-off in attention.}
Others papers talked about this trade-off, but for GNN (\citealt{Giraldo_2023}, \citealt{shao2023unifyingoversmoothingoversquashinggraph}), while this should also hold for transformers \citep{joshi2025transformersgraphneuralnetworks}, we propose below an easier way to derive it for attention :  \\
  \\
From \citealt{wu2024roleattentionmaskslayernorm} we know that the rank-collapse rates in attention depends on the attention mask graph radius (larger radius meaning slower convergence), and from \citealt{gharan2012universalupperboundgraph} we have this inequality between the radius and the first non-negative spectral value :
$$rad(G) \leq diam(G) \leq \frac{2\times48 \log(n)}{\lambda_2}$$ 
And from \citealt{topping2022understandingoversquashingbottlenecksgraphs}, we know that over-squashing depends on the first non-negative spectral value of the graph $\lambda_2$ (lower spectral value meaning higher chance of over-squashing).  \\
\\
Because $rad(G) \leq \frac{2\times48 \log(n)}{\lambda_2} \quad (n \text{: number of nodes})$, when reaching the boundary imposed by this inequality, if radius up (over-smoothing down) then spectral value down (over-squashing up) deriving the trade-off. \\ 
Do we also have the same kind of trade-off for N-simplicial attention ? Is it more or less strict ?

\subsection{Over-smoothing for N-simplicial attention} \label{subsec:ovsmN}

\cite{dong2023attentionneedpureattention} showed that attention by itself leads to the rank collapse of the tokens representation under some assumptions, by studying how its residual     
$ \text{res}(X) = X - 1 x^T, \ \text{ where } \ x = \underset{}{\text{argmin}_{u\in \mathbb{R}^d}} \| X - 1 u^T\|$
converges to zero quickly during the forward pass, for pure attention (without mask, skip and MLPs), expressing the rank collapse of the feature matrix $X$. The residual output of one self-attention layer is such that :
\begin{align*}
   \|\text{res}(\text{SA}(X))\|_{1,\infty} \leq \bigg(\frac{4 \gamma \beta }{\sqrt{d_{qk}}}\bigg) \, \|\text{res}(X) \|^{3}_{1,\infty}
\end{align*}
\noindent
With $\|W_{QK}^l\|_1 \|W_{V}^{l}\|_{1,\infty} \leq \beta$ and $\gamma$ a term that depends on the attention entries. \\ This amounts to cubic convergence to a rank-1 matrix, which becomes doubly exponential when stacking layers. For the formal theorem and proof for attention, we refer the reader to \cite{dong2023attentionneedpureattention}. 

\textbf{Do we have the same kind of result for N-simplicial attention ?}
\begin{tcolorbox}[enhanced, colback=red!10!white, colframe=red!80!black]
\begin{restatable}{theorem}{mainthm}\label{thm:main}
For the unmasked N-simplicial attention layer, we can show the following :
    \begin{equation*}
\left\| \text{res}(X') \right\|_{1,\infty} \leq \frac{4\gamma}{\sqrt{d}} \beta' \,\left\| X \right\|_{1,\infty}^{2(N-1)} \, \,\left\| \text{res}(X)   \right\|_{1,\infty}^{3}
    \end{equation*}
with $X'$ the output of the pure N-simplicial attention (without skip connection), and : \\
$$\beta' = \Big(2 \, \underset{0\leq m\leq N}{\max}\| W_K^{(m)} \|_{1,\infty}\Big)^{N+1}  \,\Big(2\,\underset{1\leq m\leq N}{\max}\| W_V^{(m)} \|_{1,\infty} \Big)^N $$
\end{restatable}
\end{tcolorbox}
\noindent
We assume infinite internal precision, making it finite could speed up convergence by grouping close features. \\
see the proof of this theorem in \autoref{appx:proof}. 

\begin{corollary}
 Extended to H heads, stacking L layers, for N-simplicial attention :\begin{align*} 
&\boxed{\| \text{res}(X^{(L)}) \|_{1,\infty} \leq \left\| \text{res}(X)   \right\|_{1,\infty}^{3^L} \Bigg(\frac{4\gamma H}{\sqrt{d}} \,\beta\,\Big(\underset{0\leq t \leq L}{\max}\| X^{(t)} \|_{1,\infty}\Big)^{2(N-1)}\Bigg)^{\frac{3^L -1}{2}} }\\
& \text{with }\beta =\Big(2 \underset{\substack{0 \le m \leq N \\ 0\leq t \leq L}}{\max}\left\| W_K^{m}(t) \right\|_{1,\infty} \Big)^{N+1} \Big(2 \underset{\substack{0 \le m \leq N \\ 0\leq t \leq L}}{\max} \left\| W_V^{m}(t) \right\|_{1,\infty} \Big)^N 
    \end{align*}
\end{corollary}
\noindent
Which gives the same results as \cite{dong2023attentionneedpureattention} for attention, case N=1 (up to a multiplicative constant) 

\subsubsection{Over-smoothing for masked N-simplicial attention}
Our previous \autoref{thm:main} relies a lot on the shift-invariance of the softmax, this causes an issue when considering the behavior of \textbf{masked} N-simplicial attention (\citealt{wu2024roleattentionmaskslayernorm} pointed this out in the case of attention (N=1), for the work of \cite{dong2023attentionneedpureattention}, which our previous theorem generalizes). 
  
  \noindent 
Masked N-simplicial attention defines a N-uniform hypergraph, whose nodes are the tokens and whose hyper-edges are the non-masked N-simplexes. To consider the over-smoothing behavior of N-simplicial attention in LLM.
We need to expand the previous result from complete to quasi-strongly connected N-uniform hypergraph (which simplicial causal mask are special cases of).


\begin{definition}[Quasi-strongly Connected]\label{def:qsc}
a directed hypergraph is said to be quasi-strongly connected if it has at least one center node (a node from which every other node in the hypergraph is reachable).
\end{definition}
\noindent
\textbf{Assumptions.} \textit{$X^{(t)}$ and $\| W_K^{(i)}(t) \|_{2}$ stay bounded for $t\geq0$, $N\geq i\geq 0$. $\|X^{(t)}\|_{1,\infty}^{N-1} \underset{ i\leq N,t\geq0}{\max} \|W_V^{(i)}(t) \|_{1,\infty}^N \leq 1 $ \\
$\mathcal{G}$ contains at least one self-loop for its center node. (i.e there is a simplex with $c$ as target and part of the sources).  }

\begin{tcolorbox}[enhanced, colback=red!10!white, colframe=red!80!black]
\begin{restatable}{theorem}{radthm}\label{thm:radius}
For the \underline{masked} N-simplicial attention, if $\mathcal{G}$ is a quasi-strongly connected hypergraph (directed or not), then $\mathcal{A}^{(t)}_{k_0 \cdots k_N}> 0 \,\, \forall(k_0 \cdots k_N) \in E(\mathcal{G)}, \forall t\geq0$. And there exists $C\geq0, \varepsilon>0$ such that at layer $t$ 
\begin{equation*}
    \| \text{res}(X^{(t)}) \|_{1,\infty} \leq C (1- \varepsilon^r )^{t/r}
\end{equation*}
for $r$ the radius of $\mathcal{G}$, expressing an exponential convergence of the tokens to a common vector (rank collapse). \\
(under some assumptions listed above).
\end{restatable}
\end{tcolorbox}\noindent
see the proof of this theorem in \autoref{appx:radius}. 
\noindent
\begin{definition}[Radius]\label{def:radius}
The radius of a quasi-strongly connected hypergraph is defined as the longest distance from a center node to any node in the hypergraph. If multiple candidates, then it is the smallest value among them.
\end{definition}

\noindent Those two over-smoothing results imply the convergence when the initial token features and parameters matrices are "small", but \cite{alman2023capturehigherordercorrelationsgeneralizing} and \cite{liang2024tensorattentiontrainingprovably} proved that for higher-order attention, having "small" token features and parameters matrices is the only way to reduce the cubic (or higher) complexity when computing for all the tokens. We have another trade-off: over-smoothing / computation complexity. 
$\newline$ $\newline$
\textbf{Continuous time. } Our previous \autoref{thm:main} and \autoref{thm:radius} extend \cite{dong2023attentionneedpureattention} and \cite{wu2024roleattentionmaskslayernorm} to N-simplicial attention in the practical case of a discrete finite amount of layers. We conjecture here that similar results of features convergence should hold for the continuous-time model of N-simplicial transformer, as it has been done for attention (\citealt{geshkovski2024emergenceclustersselfattentiondynamics}, \citealt{karagodin2024clusteringcausalattentionmasking}, 
\citealt{abella2025asymptoticbehaviorattentiontransformers}
\citealt{cimetière2025localmaxdynamicsattentiontransformers}) 
following the \textit{continuous-time over-smoothing} definition of \cite{rusch2023surveyoversmoothinggraphneural}.
$\newline$ $\newline$
\textbf{Consequences. } The purpose of this study of over-smoothing in N-simplicial attention is to allow the transfer of the knowledge and solutions gained from extensive empirical use of attention based architectures. As we have shown the similarity in their behaviors, this strongly imply the need to use residual connections \citep{dong2023attentionneedpureattention}, layer-normalization, sparse attention \citep{wu2024roleattentionmaskslayernorm} and others solutions found for attention.

\newpage
\subsection{Upper bound of N-simplicial attention Lipschitz Constant} \label{subsec:upper}
After studying the over-smoothing contraction of the tokens into the same representation, we propose to now take a look at how higher-order attention contracts two relatively close outputs, through its Lipschitz continuity.  \\
N-simplicial attention is not globally Lipschitz continuous, as it can be reduced to attention (see \autoref{subsec:reduce}) which itself is not globally Lipschitz continuous, as shown by \cite{kim2021lipschitzconstantselfattention}. We will then restrict to input $X = (x_1,\cdots,x_n) \in B^n_R 
\subseteq \mathbb{R}^{n\times d}$, where $B_R$ is the closed ball of radius R, centered on $0$.
\begin{definition}[Lipschitz constant of $f$ on $\mathcal{X}$] for  a connected subset $\mathcal{X}$ of $ \mathbb{R}^{n\times d}$. $\quad \text{Lip}(f_{|\mathcal{X}}) =  {\sup}_{X \in \mathcal{X}} \| D_X f\|_2$ \\ with $D_X f$ the differential of f on the token feature matrix input $X \in \mathbb{R}^{n\times d}$ (\citealt{federer2014geometric}, \citealt{castin2024smoothattention}).
\end{definition}
\begin{tcolorbox}[enhanced, colback=red!10!white, colframe=red!80!black]
\begin{restatable}{theorem}{lipthm}\label{thm:lips}Unmasked N-simplicial attention $f: \mathbb{R}^{n\times d} \to \mathbb{R}^{n\times d}$ is Lipschitz continuous on $B^n_R$ \\ (for $R\ge0$, and $n$ the token count), with the following local Lipschitz constant upper bound :
$$
    \text{Lip}(f_{|B^n_R}) \leq n\sqrt{2n} \,N\,V^N\,R^{N-1}\big(1 + d\,N^2 (KR)^{2(N+1)}\big)^{1/2}
$$
for $X\in B^n_R$ $\qquad V = {\max}_{1 \leq m \leq N} \| W_V^{(m)} \|_2$ $\qquad K = {\max}_{0 \leq m \leq N} \| W_K^{(m)} \|_2$ 
\end{restatable}
\end{tcolorbox} \noindent
see the proof of this theorem in \autoref{appx:lips}. 
\begin{remark}
Our previous theorem is an extension to higher-order attention of \cite{castin2024smoothattention} which proved that 
the unmasked self-attention function \( f \) with parameters \( (A, V) \) is Lipschitz continuous on \( B_n^R \), with :
\[
\text{Lip}(f|_{B_R^n}) \leq \sqrt{3} \|V\|_2 \big( \|A\|_2^2\, R^4 (4n + 1) + n \big)^{1/2}
\]
Note that \autoref{thm:lips} isn't tight in $n$ for attention, as some expressions don't simplify as nicely in the general case.\end{remark}
\noindent From \autoref{thm:lips}, we see that the dependency in $n$ (number of tokens) of the Lipschitz constant does not depend on the simplicial order $N$, this is mostly due to the softmax normalizing the different interactions and it is implying that under proper control of $R$ through normalization (e.g. LayerNorm: \citealt{ba2016layernormalization} and RMSNorm: \citealt{zhang2019rootmeansquarelayer}) the smoothing and contracting behavior of N-simplicial attention should mainly depend on the norms of the parameters matrices, with sensitivity increasing with N. $\newline \newline$
\noindent We also want to highlight that there is a difference of behavior when $R\to0$ between attention and the rest of higher-order N-simplicial attention. Indeed, there is an independent term in $R$ only for attention, which mean that we can have N-simplicial attention of arbitrary small Lipschitz constant under proper sub-normalization of the inputs feature matrix $X$. This can help reducing the model's vulnerability to adversarial inputs.

%% file: tex/part2.tex
\section{Conclusion}
This works presents a generalization of attention to higher-order interactions between tokens (\textit{N-simplicial attention}), where the increasing complexity is managed by dynamically selecting which simplexes to be kept via a light-weight, trainable router (\textit{Simplicial Path Sparse attention}), capable of respecting the causality constraint if needed. To allow for in-place replacement, we also offer a RoPE compatible variant of N-simplicial attention. We prove that despite opening to higher-order interactions, N-simplicial attention also suffers from over-smoothing (with and without mask) under some assumptions on the features and parameters weights sizes. And we derive an upper-bound of the Lipschitz constant of N-simplicial attention for bounded inputs, highlighting a difference in behavior between attention and its higher-order counterpart.
\subsection*{Acknowledgments}
This work was supported by a research grant from École des Ponts (IP Paris). We also wish to  thank Francesco Pappone, Dante Campregher, Simone Antonelli and Flavio Sartori for the helpful conversations we had.

%% file: tex/annex2.tex
\newpage
\section{Proof of \autoref{thm:main} : Over-smoothing without mask}\label{appx:proof}

\begin{tcolorbox}[enhanced, colback=red!10!white, colframe=red!80!black]
\mainthm*
\end{tcolorbox}

\noindent \textbf{Proof.}
note $ R =  \text{res}(X) = X - \mathbf{1} x^T, \ \text{ where } \ x = \underset{}{\text{argmin}_u} \| X - \mathbf{1} u^T\| $ \hfill then $ X = \mathbf{1}x^\top + R $ \\
In the N-simplicial attention logits tensor $\mathcal{L}$ each coefficient is of form,
\begin{align*}
    \mathcal{L}_{k_0\cdots k_N} \,
&= \,\,\, \sum_{a=1}^d \,\prod_{i=0}^{N} 
\big( X W_{K}^{(i)} \big)_{k_i a} \\
&= \,\,\,\sum_{a=1}^d \, \prod_{i=0}^{N} 
\big( (\mathbf{1}x^T + R)  W_{K}^{(i)} \big)_{k_i a} \\
&= \,\,\, 
\sum_{a=1}^d \,
\sum_{S\subseteq\{0,\dots,N\}}
\left(
\prod_{i\in S} (RW_{K}^{(i)})_{k_i a}
\right)
\left(
\prod_{j\notin S} (\mathbf{1}x^TW_{K}^{(j)})_{k_ja}
\right) \\
\end{align*}
We now separate the terms by $|S|$, the number of residual factors kept, with $C:\{|S|=0 \}$, $\widetilde{\mathcal{L}} : \{|S|=1 \}$ (all first-order residual contributions) and $\mathcal{E} : \{|S| \geq 2\}$ (all higher-order residual contributions).
\[
{\mathcal{C}}_{k_0\cdots k_N}
:= 
\sum_{a=1}^d \,
\sum_{\substack{S\subseteq\{0,\dots,N\}\\ |S| =0}}
\left(
\prod_{i\in S} (RW_{K}^{(i)})_{k_i a}
\right)
\left(
\prod_{j\notin S} (\mathbf{1}x^TW_{K}^{(j)})_{k_ja}
\right)
= \sum_{a=1}^d \, \prod_{j=0}^N (\mathbf{1}x^\top W_{K}^{(j)})_{k_j a}.
\]
noting that for  $\forall k_j$, 
$
(\mathbf{1}x^\top W_{K}^{(j)})_{k_j a}
= \sum_{b=1}^1 1_{k_j b}\,(x^\top W_{K}^{(j)})_a
= (x^\top W_{K}^{(j)})_a,
$ is independent of $k_j$, the product over \(j\) does not depend on the tuple \((k_0,\dots,k_N)\), so $C$ is a constant tensor.

\[
\widetilde{\mathcal{L}}_{k_0\cdots k_N}
:= 
\sum_{a=1}^d \,
\sum_{\substack{S\subseteq\{0,\dots,N\}\\ |S|=1}}
\left(
\prod_{i\in S} (RW_{K}^{(i)})_{k_i a}
\right)
\left(
\prod_{j\notin S} (\mathbf{1}x^TW_{K}^{(j)})_{k_ja}
\right)
\]

\[
{\mathcal{E}}_{k_0\cdots k_N}
:= 
\sum_{a=1}^d \,
\sum_{\substack{S\subseteq\{0,\dots,N\}\\ |S| \geq2}}
\left(
\prod_{i\in S} (RW_{K}^{(i)})_{k_i a}
\right)
\left(
\prod_{j\notin S} (\mathbf{1}x^TW_{K}^{(j)})_{k_ja}
\right)
\]
Hence the logits decompose as
$ \boxed{
\mathcal{L} = C  +\widetilde{\mathcal{L}} + \mathcal{E}
} $ \\
And because softmax has a shift invariance property, the attention score are :
\[
 \mathcal{A} := \mathrm{softmax}(\mathcal{L})  = \mathrm{softmax}(C  +\widetilde{\mathcal{L}} + \mathcal{E}) \underset{\text{invariance}}{=} \mathrm{softmax}(\widetilde{\mathcal{L}} + \mathcal{E})
\]
\noindent
by defining $\forall i\leq N, \,\,e^{(i)} \in \mathbb{R}^{n}$ such that $e^{(i)}_{m} := \overset{d}{\underset{a=1}{\sum}} \, (RW_K^{(i)})_{ma} \, \underset{j \neq i}{\prod} (x^TW_K^{(j)})_a \,$  we have $
\widetilde{\mathcal{L}}_{k_0\cdots k_N} = \overset{N}{\underset{i=0}{\sum}} e^{(i)}_{k_i} $ \\ \noindent
i.e  
$
\widetilde{\mathcal{L}}
= \overset{N}{\underset{i=0}{\sum}}
\Big(
\mathbf{1}_{k_0} \otimes \cdots \otimes \mathbf{1}_{k_{i-1}} 
\otimes e^{(i)} \otimes 
\mathbf{1}_{k_{i+1}} \otimes \cdots \otimes \mathbf{1}_{k_N} 
\Big),
$
where \(\mathbf{1}_{k_j} \in \mathbb{R}^{n}\) are vector of ones, (see \cref{lem:sum}).
Here, each term in the sum depends on only one axis (the one corresponding to $e^{(i)}$) and is constant along the others.
Each term is "rank-1 along all axes" (separable tensor).
This will be useful to show that the result of the contraction with the non-residual part of the $V_m$ matrix will be of rank-1 and row-constant.

\noindent

\begin{tcolorbox}
\begin{lemma*}[Lemma A.3, \citealt{dong2023attentionneedpureattention}]
Suppose that \(P\) is the row-stochastic matrix associated with \(A\), and let 
\(\widetilde{P}\) be the one associated with 
\(\widetilde{A} = A - E\) 
for some matrix \(E\) satisfying 
\(|E_{ij} - E_{ij'}| \le 1.256\) 
for every \(i, j, j'\).
Then, we have the entry-wise inequality
\[
(\mathbf{I} - D)\,\widetilde{P} \;\le\; P \;\le\; (\mathbf{I} + 2D)\,\widetilde{P},
\]
where \(D\) diagonal matrix,
$
D_{ii} = \max_{j,j'} \big| \delta_i^\top E (\delta_j - \delta_{j'}) \big|,
$
and the inequality is understood entry-wise.
\end{lemma*}
\end{tcolorbox}
\noindent
The Lemma A.3 of Y.Dong et al. paper still work when $\mathcal{A} \in \mathbb{R}^{n \times m} $ is a rectangle matrix (and not a square matrix), so applied to the flattened attention tensor (here with $m = n^N$, all the other dimensions but the first) with $P = \mathrm{softmax}(\widetilde{\mathcal{L}} + \mathcal{E})$ and $ \tilde{P} = \mathrm{softmax}(\widetilde{\mathcal{L}})$ \\

\noindent
Let's define $\Delta = \mathrm{softmax}(\widetilde{\mathcal{L}} + \mathcal{E}) - \mathrm{softmax}(\widetilde{\mathcal{L}}) = P - \tilde{P}$ \\
By the lemma A.3, if the flattened, $ \mathcal{E}$ satisfies $|\mathcal{E}_{ij} - \mathcal{E}_{ij'}| \le 1.256 $ for every $i, j, j'$. \\ then $ | \Delta| = | P - \tilde{P}| \leq 2 D |\tilde{P}| $ (entry-wise inequality) with $D$ diagonal matrix of the lemma, non-negative terms. \\

\noindent
The output of a N-simplicial attention layer (without skip connection), is then :
\begin{align*}
X'_{k_{0}d} &=   \sum_{k_1, \ldots,k_N}  \mathcal{A}_{k_0\cdots k_N} \, \,\cdot\, \, \prod_{m=1}^{N} \Big( X W_{V}^{(m)} \Big)_{k_{m}d} \\
&= \frac{1}{\sqrt{d}} \sum_{k_1, \ldots,k_N}  \big(\text{softmax}(\widetilde{\mathcal{L}}) + \Delta \big)_{k_0\cdots k_N} \, \,\cdot\, \, \prod_{m=1}^{N} \Big( X W_{V}^{(m)} \Big)_{k_{m}d} \\
&= \frac{1}{\sqrt{d}}\sum_{k_1, \ldots,k_N}  \big(\text{softmax}(\widetilde{\mathcal{L}})\big)_{k_0\cdots k_N} \cdot \prod_{m=1}^{N} \Big( X W_{V}^{(m)} \Big)_{k_{m}d} 
\quad+\,\,\,\,\,\, \frac{1}{\sqrt{d}}
\sum_{k_1, \ldots,k_N}  (\Delta)_{k_0\cdots k_N} \cdot\prod_{m=1}^{N} \Big( X W_{V}^{(m)} \Big)_{k_{m}d}
\end{align*}

\noindent
And $\sum_{k_1, \ldots,k_N}  \big(\text{softmax}(\widetilde{\mathcal{L}})\big)_{k_0\cdots k_N} \, \,\cdot\, \, \prod_{m=1}^{N} \Big( X W_{V}^{(m)} \Big)_{k_{m}\,d}$ is row-constant (see \cref{lem:tilde}) \\ can thus be used to bound the residual of $X'$. (even without separating the $XW_V$). 
\begin{align*}
    X'_{k_{0}d} &=   \sum_{k_1, \ldots,k_N}  \mathcal{A}_{k_0\cdots k_N} \, \,\cdot\, \, \prod_{m=1}^{N} \Big( X W_{V}^{(m)} \Big)_{k_{m}d} \\
    &=   \sum_{k_1, \ldots,k_N}  \mathcal{A}_{k_0\cdots k_N} \, \,\cdot\, \, \prod_{m=1}^{N} \Big( (\mathbf{1}x^T + R)  W_{V}^{(m)} \Big)_{k_{m}d} \\
    &= \sum_{k_1, \ldots,k_N}  \mathcal{A}_{k_0\cdots k_N} \,
\sum_{S\subseteq\{0,\dots,N\}}
\left(
\prod_{i\in S} (RW_{V}^{(i)})_{k_{i}d}
\right)
\left(
\prod_{j\notin S} (\mathbf{1}x^TW_{V}^{(j)})_{k_{j}d}
\right)  \\
 &= \sum_{k_1, \ldots,k_N}  \mathcal{A}_{k_0\cdots k_N}
\prod_{j=0}^{N} (\mathbf{1}x^TW_{V}^{(j)})_{k_{j}d}
\quad + \quad
\sum_{k_1, \ldots,k_N}  \mathcal{A}_{k_0\cdots k_N} 
\sum_{\substack{S\subseteq\{0,\dots,N\}\\ |S|\geq 1}}
\big(
\prod_{i\in S} (RW_{V}^{(i)})_{k_{i}d}
\big)
\Big(
\prod_{j\notin S} (\mathbf{1}x^TW_{V}^{(j)})_{k_{j}d}
\Big)
\end{align*}\noindent
Because $\mathbf{1}x^TW_{V}^{(j)}$ is row-constant, and $ \mathcal{A}$ is row-stochastic, summing over \(k_1,\ldots,k_N\) just yields 1, so:
\[
\sum_{k_1,\ldots,k_N}
\mathcal{A}_{k_0\cdots k_N}
\prod_{j=0}^{N} 
(\mathbf{1}x^T W_V^{(j)})_{k_j d}
=
\prod_{j=0}^{N} 
(x^T W_V^{(j)})_d.
\]
The whole first term above (in $X'$) is row-constant, it will disappear bounding the residual of $X'$. (also contains a bit of $\text{softmax}(\widetilde{\mathcal{L}})$ row-constant terms, below is the fusion keeping in mind the shared part). \\
\noindent
Combining the two approaches above, the row-constant terms are :
\begin{equation*}
    u = \sum_{k_1,\ldots,k_N}
\mathcal{A}_{k_0\cdots k_N}
\prod_{j=0}^{N} 
(\mathbf{1}x^T W_V^{(j)})_{k_j d} \quad+\frac{1}{\sqrt{d}} \sum_{k_1, \ldots,k_N} \text{softmax}(\widetilde{\mathcal{L}})_{k_0\cdots k_N} \cdot \sum_{\substack{S\subseteq\{0,\dots,N\}\\ |S|\geq 1}}
\prod_{i\in S} (RW_{V}^{(i)})_{k_{i}d}
\prod_{j\notin S} (\mathbf{1}x^TW_{V}^{(j)})_{k_{j}d}
\end{equation*}
\noindent
And by definition of the residual, we have :
$   \left\| \text{res}(X') \right\| \leq \left\| X' - u
\right\| $ i.e 
\begin{equation*}
    \left\| \text{res}(X') \right\|_{1,\infty} \leq \left\| X' - u
\right\|_{1,\infty}
=
\Bigg\| \frac{1}{\sqrt{d}}
\sum_{k_1, \ldots,k_N}  (\Delta)_{k_0\cdots k_N} \cdot \sum_{\substack{S\subseteq\{0,\dots,N\}\\ |S|\geq 1}}
\prod_{i\in S} (RW_{V}^{(i)})_{k_{i}d}
\prod_{j\notin S} (\mathbf{1}x^TW_{V}^{(j)})_{k_{j}d} \Bigg\|_{1,\infty}
\end{equation*}
\noindent 
The main issue is that there's still some terms with $x^T$ inside, they each have at least one $RW_V$, but we need to control the rest. But the norm of X is bounded anyway, and $\left\| R \right\| = \left\| \text{res}(X) \right\|  \leq \left\| X \right\|$ and $\left\| \mathbf{1}x^T \right\| \leq \left\| X \right\|$
\begin{align*}
    \| \text{res}(X') \|_{1,\infty} &\leq 
\Bigg\| \frac{1}{\sqrt{d}}
\sum_{k_1, \ldots,k_N}  (\Delta)_{k_0\cdots k_N} \cdot \sum_{\substack{S\subseteq\{0,\dots,N\}\\ |S|\geq 1}}
\prod_{i\in S} (RW_{V}^{(i)})_{k_{i}d}
\prod_{j\notin S} (\mathbf{1}x^TW_{V}^{(j)})_{k_{j}d} \Bigg\|_{1,\infty}  \\
& \leq \frac{1}{\sqrt{d}}\left\| \Delta \right\|_{1,\infty} \left\| R \right\|_{1,\infty} \Big(\prod_{1 \leq m\leq N}\left\| W_V^{m} \right\|_{1,\infty} \Big) \left\| X \right\|_{1,\infty}^{N-1} \, 2^{N}
\end{align*}
\noindent
And for the norm of $\Delta$, we have :
$
    \left\| \Delta \right\|_{1,\infty} \leq 2 \left\| D \right\|_{1,\infty}\left\| \tilde{P} \right\|_{1,\infty} \leq 2 \left\| D \right\|_{1,\infty}
$
\noindent
By definition of $D$,  we have that 
\begin{align*}
    \|D\|_{1,\infty}^{2} 
    &= \|D\|_1 \|D \|_\infty \\
    &= \max_{i,j,j'} |\mathcal{E}_{ij} - \mathcal{E}_{ij'} | \,\sum_{i} \max_{j,j'} |\mathcal{E}_{ij} - \mathcal{E}_{ij'}| \\
    &= \max_{i,j,j'}|A_{ij}-A_{ij'}|  \sum_{i} \max_{j,j'} |A_{ij} - A_{ij'} | \\    
    &\leq \gamma^2 \left(\max_{j,j'} \sum_{i} |A_{ij} - A_{ij'} |\right)^2 \tag{by assumption}
    = \gamma^2 \left(\max_{j,j'} \sum_{i} |E_{ij} - E_{ij'} |\right)^2 \\
    &\leq 4 \gamma^2 \, \|\mathcal{E}\|_{1}^2 \\
\end{align*}
\noindent And, we have $
{\mathcal{E}}_{k_0\cdots k_N}
= 
\underset{a=1}{\overset{d}{\sum}} \,
\underset{\substack{S\subseteq\{0,\dots,N\}\\ |S| \geq2}}{\sum}
\left(
\underset{i \in S}{\prod} (RW_{K}^{(i)})_{k_i a}
\right)
\left(
\underset{j \notin S}{\prod} (\mathbf{1}x^TW_{K}^{(j)})_{k_ja}
\right)
$ \\
so $\|\mathcal{E}\|_{1} \leq \|R\|_{1,\infty}^{2} \, (2^{N+1}-(N+2)) \,\Big(\underset{0 \leq m\leq N}{\prod}\left\| W_K^{m} \right\|_{1,\infty} \Big) \|X\|_{1,\infty}^{N-1} $
$\qquad$ this gives
%
\begin{align*}
    \left\| \text{res}(X') \right\|_{1,\infty}
&\leq \frac{4\gamma}{\sqrt{d}} \|R\|_{1,\infty}^{2}\Big(2\underset{0\leq m\leq N}{\max}\left\| W_K^{m} \right\|_{1,\infty} \Big)^{N+1} \|X\|_{1,\infty}^{N-1} \, \quad  \left\| R \right\|_{1,\infty} \Big(2 \max_{1\leq m\leq N}\left\| W_V^{m} \right\|_{1,\infty} \Big)^N \left\| X \right\|_{1,\infty}^{N-1}  \\
\left\| \text{res}(X') \right\|_{1,\infty}&\leq \frac{4 \gamma}{\sqrt{d}} \Big(2\underset{0\leq m\leq N}{\max}\left\| W_K^{m} \right\|_{1,\infty} \Big)^{N+1}\Big(2 \,\max_{m\leq N}\left\| W_V^{m} \right\|_{1,\infty} \Big)^N \left\| X \right\|_{1,\infty}^{2(N-1)}  \,\left\| \text{res}(X)   \right\|_{1,\infty}^{3}
\end{align*}
Gives mostly same results for attention as \cite{dong2023attentionneedpureattention}, bit worse as went from $(2^{N+1}-(N+2))$ to $2^{N+1}$.
\begin{equation*}
\left\| \text{res}(X') \right\|_{1,\infty} \leq \frac{4 \gamma}{\sqrt{d}} \Big(2\underset{0\leq m\leq N}{\max}\left\| W_K^{m} \right\|_{1,\infty} \Big)^{N+1}\Big(2 \,\max_{m\leq N}\left\| W_V^{m} \right\|_{1,\infty} \Big)^N \left\| X \right\|_{1,\infty}^{2(N-1)}  \,\left\| \text{res}(X)   \right\|_{1,\infty}^{3}
\end{equation*}
\noindent Extended to H attention-heads, and stacking L layers, for N-simplicial attention : \begin{equation*}
\left\| \text{res}(X^{L}) \right\|_{1,\infty} \leq \left\| \text{res}(X)   \right\|_{1,\infty}^{3^L} \Bigg(\frac{4\gamma H}{\sqrt{d}} \Big(2 \underset{0\leq m\leq N}{\max}\left\| W_K^{m} \right\|_{1,\infty} \Big)^{N+1} \Big(2\max_{1\leq m\leq N}\left\| W_V^{m} \right\|_{1,\infty} \Big)^N \Big(\underset{0\leq t \leq N}{\max}\left\| X^t \right\|_{1,\infty}\Big)^{2(N-1)}\Bigg)^{\frac{3^L -1}{2}}
    \end{equation*}
 \qed

\begin{lemma} \label[lemma]{lem:sum}
we have $
\widetilde{\mathcal{L}}_{k_0\cdots k_N} = \overset{N}{\underset{i=0}{\sum}} e^{(i)}_{k_i} $ also writable as :
$
\widetilde{\mathcal{L}}
= \overset{N}{\underset{i=0}{\sum}} 
\Big(
\mathbf{1}_{k_0} \otimes \cdots \otimes \mathbf{1}_{k_{i-1}} 
\otimes e^{(i)} \otimes 
\mathbf{1}_{k_{i+1}} \otimes \cdots \otimes \mathbf{1}_{k_N} 
\Big),
$
With $\forall i\leq N, \,\,e^{(i)} \in \mathbb{R}^{n}$ such that $e^{(i)}_{m} := \overset{d}{\underset{a=1}{\sum}} \, (RW_K^{(i)})_{ma} \, \underset{j \neq i}{\prod} (x^TW_K^{(j)})_a \,$ and \(\mathbf{1}_{k_j} \in \mathbb{R}^{n}\) vectors of ones
\end{lemma} \noindent \textbf{Proof.}
\begin{align*}
    \widetilde{\mathcal{L}}_{k_0\cdots k_N} &= 
    \sum_{a=1}^d \, \sum_{\substack{S\subseteq\{0,\dots,N\}\\ |S|=1}}
    \left( \prod_{i\in S} (RW_{K}^{(i)})_{k_i a} \right)
    \left( \prod_{j\notin S} (\mathbf{1}x^TW_{K}^{(j)})_{k_ja} \right) \\
    &= \sum_{i=0}^{N} \,\sum_{a=1}^d \, (RW_K^{(i)})_{k_{i}a} \, \prod_{j \neq i}(\mathbf{1}x^T W_K^{(j)})_{k_ja} \\ \intertext{and  $\, \forall j, \, (\mathbf{1}x^T W_K^{(j)})_{k_ja} = (x^TW_K^{(j)})_a$  is independent of row index $k_j$, (because of $\mathbf{1}x^T$)} 
    &= \sum_{i=0}^{N} \,\sum_{a=1}^d \, (RW_K^{(i)})_{k_{i}a} \, \prod_{j \neq i}(x^TW_K^{(j)})_a
\end{align*} \noindent
by defining $\forall i\leq N, \,\,e^{(i)} \in \mathbb{R}^{n}$ such that $e^{(i)}_{m} := \overset{d}{\underset{a=1}{\sum}} \, (RW_K^{(i)})_{ma} \, \underset{j \neq i}{\prod} (x^TW_K^{(j)})_a \,$  we have $
\widetilde{\mathcal{L}}_{k_0\cdots k_N} = \overset{N}{\underset{i=0}{\sum}} e^{(i)}_{k_i} $ \\
\noindent Written for the whole tensor at once, as a tensor decomposition, where \(\mathbf{1}_{k_j} \in \mathbb{R}^{n}\) are vector of ones, we have : 
\[
\widetilde{\mathcal{L}}
= \sum_{i=0}^{N} 
\Big(
\mathbf{1}_{k_0} \otimes \cdots \otimes \mathbf{1}_{k_{i-1}} 
\otimes e^{(i)} \otimes 
\mathbf{1}_{k_{i+1}} \otimes \cdots \otimes \mathbf{1}_{k_N} 
\Big),
\] \\
\noindent
\begin{lemma} \label[lemma]{lem:tilde}
    $\sum_{k_1, \ldots,k_N}  \big(\text{softmax}(\widetilde{\mathcal{L}})\big)_{k_0\cdots k_N} \, \,\cdot\, \, \prod_{m=1}^{N} \Big( X W_{V}^{(m)} \Big)_{k_{i}\,d}$ is row-constant.
\end{lemma} \noindent
\textbf{Proof.} Using \cref{lem:sum}, we have
$ \quad 
\widetilde{\mathcal{L}}_{k_0 k_1\cdots k_N}
= \overset{N}{\underset{i=0}{\sum}} e^{(i)}_{k_i}
$ \\
When applying softmax we treat
\((k_1,\dots,k_N)\) as the softmax domain for each fixed query index \(k_0\).  The term \(e^{(0)}_{k_0}\) is independent of \((k_1,\dots,k_N)\), as a constant offset
it's canceled by the shift-invariance.
Thus, defining
$
\widehat{\mathcal{L}}_{k_1\cdots k_N} := \overset{N}{\underset{i=1}{\sum}} e^{(i)}_{k_i}
\quad $ we have for each fixed \(k_0\), $
\operatorname{softmax}\!\big(\widetilde{\mathcal{L}}_{k_0,\cdot}\big)_{k_1\cdots k_N}
= \operatorname{softmax}\!\big(\widehat{\mathcal{L}}\big)_{k_1\cdots k_N}.
\quad$ and \\
\noindent
\[
\exp\!\big(\widehat{\mathcal{L}}_{k_1\cdots k_N}\big)
= \prod_{i=1}^N \exp\!\big(e^{(i)}_{k_i}\big).
\]
Therefore the softmax itself factorizes:
\[
\operatorname{softmax}\!\big(\widehat{\mathcal{L}}\big)_{k_1\cdots k_N}
=
\frac{\prod_{i=1}^N \exp(e^{(i)}_{k_i})}
{\sum_{t_1,\dots,t_N}\prod_{i=1}^N \exp(e^{(i)}_{t_i})}
=
\prod_{i=1}^N p^{(i)}_{k_i},
\]
where
$
p^{(i)}_{k_i} \;:=\; \frac{\exp(e^{(i)}_{k_i})}{\sum_{t=1}^n \exp(e^{(i)}_{t})}
$
is the (axis-wise) softmax of \(e^{(i)}\) over its index \(k_i\). \\
\noindent
Now consider the contraction (for a fixed feature coordinate \(a\)):
\[
Y_{k_0,a}
:= \sum_{k_1,\dots,k_N} \operatorname{softmax}\!\big(\widetilde{\mathcal{L}}\big)_{k_0 k_1\cdots k_N}
\prod_{m=1}^N V^{(m)}_{k_m,a}.
\]
Using the factorization we get
\[
Y_{k_0,a}
= \sum_{k_1,\dots,k_N} \Big(\prod_{i=1}^N p^{(i)}_{k_i}\Big)\;\prod_{m=1}^N V^{(m)}_{k_m,a}
= \prod_{m=1}^N \Big(\sum_{k_m} p^{(m)}_{k_m}\, V^{(m)}_{k_m,a}\Big).
\]
Each factor on the right involves only \(p^{(m)}\) and \(V^{(m)}\) (both independent of \(k_0\)), hence the product is independent of \(k_0\). Consequently \(Y_{k_0,a}\) is the same for every \(k_0\) : row-constant  
\qed

%% file: tex/annex3.tex
\newpage
\section{Proof of \autoref{thm:radius} :
Over-smoothing for \underline{masked} N-simplicial attention}
\label{appx:radius} 
\begin{tcolorbox}[enhanced, colback=red!10!white, colframe=red!80!black]
\radthm*
\end{tcolorbox}

\noindent \textbf{Assumptions.} $X^{(t)}$ and $\Big(\| W_K^{(i)}(t) \|_{2}\Big)_{0\leq i\leq N}$ stay bounded for $t\geq0$. $\mathcal{G}$ contains self-loops for its center nodes. \\ \noindent (a self-loop here is for token $i$ there is a N-simplex where $i$ is both the target and part of the sources).  
\\  \\
\noindent \textbf{Proof.}
In the \underline{masked} N-simplicial attention logits tensor $\mathcal{L}$ each coefficient is of form,
\begin{equation*}
    \mathcal{L}^{(t+1)}_{k_0\cdots k_N} \,
= \,\,\, \sum_{a=1}^d \,\prod_{i=0}^{N} 
\big( X^{(t)} W_{K}^{(i)} \big)_{k_i a}
\end{equation*}
\begin{equation*}
\mathcal{A}^{(t+1)}_{k_0 \cdots k_N} =
\begin{cases}
\displaystyle \text{softmax}(\mathcal{L}^{(t+1)}_{k_0\cdots k_N}) & \text{if $(k_0, \ldots, k_N) \in E(\mathcal{G})$} \\
\quad \quad 0 & \text{if not.}
\end{cases}
\end{equation*}
\noindent
As $X^{(t)}$ trajectory stay bounded, and as there exists a constant $b \in \mathbb{R}$ such that for all $t\geq0$, we have $\underset{0\leq i \leq N}{\max}\Big(\| W_K^{(i)}(t) \|_{2}\Big) \leq b$ (see Assumptions, verified in practice), then there exists a constant $C$ such that \begin{equation*}
    - C^2 \leq \mathcal{L}^{(t)}_{k_0\cdots k_N} \,
\leq C^2
\end{equation*}
This implies (due to the softmax) that there exists $\varepsilon \geq 0$ such that $\mathcal{A}_{k_0 \cdots k_N} \geq \varepsilon$ for all $(k_0, \ldots, k_N) \in E(\mathcal{G})$.  \\
\\
\noindent
Let's reflect the flow of information by the direction of our simplexes, in the simplex $(q=k_0 , k_1, \ldots,k_N)$, $q$ is the sole target, $\{k_i, 1\leq i\leq N\}$ are the sources of the directed hyperedge associated.  \\  \\ \noindent
We will assume that $\mathcal{G}$ contains self-loops for its center nodes \\
(a self-loop here is for token $i$ there is a N-simplex where $i$ is both the target and part of the sources). \\
\\
We define, $M^{(t)}_j \coloneqq \underset{i \leq n}{\max}(X^{(t)}_{ij}) $ and $m^{(t)}_j \coloneqq \underset{i \leq n, \tau \geq t}{\min}(X^{(\tau)}_{ij})-\epsilon $ (the offset of $\epsilon$ is to ensure $\rho>0$ later on). \\
\noindent
for each row $i$, there is $1 \geq \rho_i >0$, such that $X_{ij}^{(t)} = \rho_i \,m^{(t)}_j + (1-\rho_i)M^{(t)}_j$\\ Because of the self-loops assumption for $c$ center node, there is a simplex with both $c$ as a source and target,\\ let's suppose without any loss of generality that it is the $(c, \,c, k_2,\cdots, k_N)$ simplex :
\begin{align*}
    X_{cj}^{(t+1)} &\leq \mathcal{A}_{\,c \,c k_2\cdots k_N} X_{cj} \,(\|X\|_{1,\infty}^{N-1}V^N) + (1-\mathcal{A}_{\,c \,c k_2\cdots k_N}) M_j^{(t)}(\|X\|_{1,\infty}^{N-1}V^N) \\&\leq^{\star} \mathcal{A}_{\,c \,c k_2\cdots k_N} X_{cj} + (1-\mathcal{A}_{\,c \,c k_2\cdots k_N}) M_j^{(t)}
\end{align*}
$\star$ assumes that $(\|X\|_{1,\infty}^{N-1}V^N) \leq 1\quad $ which gives some new condition on $X$ and $V \coloneqq \underset{t\geq0, i\leq N}{\max} \|W_V^{(i)}(t) \|_{1,\infty}$.
\begin{align*}
    X_{cj}^{(t+1)}
    &\leq \varepsilon X_{cj}^{(t)} + (1-\varepsilon) M_j^{(t)} \\
    &\leq \varepsilon  (\rho_c \,m^{(t)}_j + (1-\rho_c)M^{(t)}_j) + (1-\varepsilon) M_j^{(t)} \\
    &=   \rho_c\,\varepsilon \,m^{(t)}_j + (1-\rho_c\,\varepsilon)M^{(t)}_j
\end{align*}
We generalise this result for $(t+T)$ in \cref{lem:center}. \\ \noindent
Let $c \in \mathcal{G}$ be the center node of the mask's graph. \\
Noting $\mathcal{V}_k(c)$ the set of nodes that are exactly k-hop neighbors of $c$, across the directed simplexes. \\
For any $i_1 \in \mathcal{V}_1(c)$, there is at least one simplex that has $i_1$ as its target and $c$ in one of the sources.
\begin{align*}
    X^{(t+1)}_{i_1\,j} &\leq \mathcal{A}_{\,c \,i_1 k_2\cdots k_N}X_{cj}^{(t)} + (1-\mathcal{A}_{\,c \,i_1 k_2\cdots k_N})M_j^{(t)} \quad \\
    &\leq \varepsilon \big(\rho_c \,m^{(t)}_j + (1-\rho_c)M^{(t)}_j\big) + (1-\varepsilon)M^{(t)}_j  \\
    &= \rho_c\, \varepsilon \,m^{(t)}_j + (1-\rho_c\,\varepsilon \,M^{(t)}_j)
\end{align*} 
\noindent
We have:$ \quad
X^{(t+T)}_{c\,j} \leq \rho_c \,\varepsilon^T m^{(t)}_j + (1-\rho_c \,\varepsilon^T)M^{(t)}_j $ from \cref{lem:center}, and thus we can adapt it :\\
(we also have $M^{(t+1)}_j \leq M^{(t)}_j (\|X\|_{1,\infty}^{N-1}V^N) \leq^{\star} M^{(t)}_j$ using the same previous $\star$ assumption).
\begin{align*}
    X^{(t+T)}_{i_1\,j} &\leq \mathcal{A}_{\,c \,i_1 k_2\cdots k_N}X_{cj}^{(t+T-1)} + (1-\mathcal{A}_{\,c \,i_1 k_2\cdots k_N})M_j^{(t+T-1)} \quad \\
    &\leq \varepsilon \big(\rho_c\, \varepsilon^{T-1}\,m^{(t)}_j + (1-\rho_c\, \varepsilon^{T-1})M^{(t)}_j\big) + (1-\varepsilon)M^{(t)}_j  \\
    &= \rho_c\, \varepsilon^{T} \,m^{(t)}_j + (1-\rho_c\,\varepsilon^T \,M^{(t)}_j)
\end{align*}
$
X^{(t+T)}_{i_1\,j} \leq \rho_c \,\varepsilon^T m^{(t)}_j + (1-\rho_c \,\varepsilon^T)M^{(t)}_j
$  \\
\\ \noindent
Idem for any $i_2 \in \mathcal{V}_2(c)$, there is at least one simplex with target $i_2$ and intermediate $i_1 \in \mathcal{V}_1(c)$ as a source.\\
we still have $M^{(t+1)}_j \leq M^{(t)}_j (\|X\|_{1,\infty}^{N-1}V^N) \leq^{\star} M^{(t)}_j$ using the same previous $\star$ assumption.
\begin{align*}
    X^{(t+2)}_{i_2\,j} &\leq \mathcal{A}_{\,i_1 \,i_2\, k_2\cdots k_N}X_{i_1j}^{(t+1)} + (1-\mathcal{A}_{\,i_1 \,i_2\, k_2\cdots k_N})M_j^{(t+1)} \\
    &\leq \varepsilon \big(\rho_c \,\varepsilon\,m^{(t)}_j + (1-\rho_c\,\varepsilon)M^{(t)}_j\big) + (1-\varepsilon)M^{(t)}_j \\
    &= \rho_c\, \varepsilon^2 \,m^{(t)}_j + (1-\rho_c\,\varepsilon^2 \,M^{(t)}_j)
\end{align*}
\noindent $
X^{(t+T)}_{i_2\,j} \leq \rho_c \,\varepsilon^T m^{(t)}_j + (1-\rho_c \,\varepsilon^T)M^{(t)}_j
$ in the same way we did for $i_1$.\\
\\ \noindent
By induction, we go through the whole graph on $r$ iteration, as $\mathcal{G}$ is quasi-strongly connected and $c$ is a center node, $r$ the radius of $\mathcal{G}$ (see \cite{wu2024roleattentionmaskslayernorm} page 15 for more details). \\
we have $\forall i\in \mathcal{G}, \quad X_{ij}^{(t+r)} \leq \rho\, \varepsilon^r m^{(t)}_j + (1-\rho\, \varepsilon^r)M^{(t)}_j$ \\
  \\
By taking the maximum, it follows : $M_{j}^{(t+r)} \leq \rho\, \varepsilon^r m^{(t)}_j + (1-\rho\, \varepsilon^r)M^{(t)}_j$ \\
Using that $\forall j\leq d, \quad m^{(t+r)}_j \geq m^{(t)}_j$, we have : 
\begin{align*}
    M_{j}^{(t+r)} - m_j^{(t+r)}&\leq \rho\, \varepsilon^r m^{(t)}_j + (1-\rho\, \varepsilon^r)M^{(t)}_j - m_j^{(t)} \\
    &= (1 -\rho \, \varepsilon^r)\big(M_j^{(t)} - m_j^{(t)} \big)
\end{align*}
\noindent
which then by induction gives us :
 $ M^{(t)}_{j} - m^{(t)}_{j} \leq C_j \,(1 - \rho\varepsilon^r)^{t/r}  $ and $\rho >0$, noting $\epsilon \coloneqq \rho^{\frac{1}{r}}\,\varepsilon > 0$\\
\noindent
$ M^{(t)}_{j} - m^{(t)}_{j} \leq C_j \,(1 - \epsilon^r)^{t/r}$ and by definition of $M$ and $m$, we have
$\big| X^{(t)}_{m,j} - X^{(t)}_{n,j} \big|^2 \leq \big( M^{(t)}_{j} - m^{(t)}_{j} \big)^2 $\\
we can now finish the proof in the same way \cite{wu2024roleattentionmaskslayernorm} did, by definition of the residual :
\begin{align*}
\left\| \text{res}(X^{(t)}) \right\|_{1,\infty} 
&\leq \left\| X^{(t)} - \frac{\underline{1}(\underline{1}^T X^{(t)})}{N} \right\|_{1,\infty} \\
&\leq \left\| X^{(t)} - \frac{\underline{1}(\underline{1}^T X^{(t)})}{N} \right\|_F = \sqrt{ \sum_{j=1}^{d} \left\| X^{(t+1)}_{:,j} - \frac{\underline{1}(\underline{1}^T X^{(t)}_{:,j})}{N} \right\|_2^2 } \\
&= \sqrt{ \frac{1}{2N} \sum_{j=1}^{d} \sum_{m=1}^N \sum_{n=1}^N \big| X^{(t)}_{m,j} - X^{(t)}_{n,j} \big|^2 } \\
&\leq \sqrt{ \frac{N}{2} \sum_{j=1}^{d} C_j^2 (1 - \varepsilon^r)^{2t/r} } \le \sqrt{C \, (1 - \varepsilon^r)^{2t/r} } \\
&\leq  C' (1 - \varepsilon^r)^{t/r}
\end{align*}

\newpage

\begin{lemma} \label[lemma]{lem:center}
For $c$ a center node, we have
$ \qquad X^{(t+T)}_{i\,j} \leq \rho_i \,\varepsilon^T m^{(t)} + (1-\rho_i \,\varepsilon^T)M^{(t)}$ \\
with $M^{(t)}_j \coloneqq \underset{i \leq n}{\max}(X^{(t)}_{ij}) \qquad$ $m^{(t)}_j \coloneqq \underset{i \leq n}{\min}(X^{(t)}_{ij})-\varepsilon\qquad \qquad$
$\rho_i \text{ such that }X_{ij} = \rho_i \,m^{(t)}_j + (1-\rho_i)M^{(t)}_j$
\end{lemma}  \noindent \textbf{Proof.}
as we did previously, assuming that $\|X\|_{1,\infty} V \leq 1$ which gives some new condition on $X, V$ ($\star$)
\begin{align*}
    X_{ij}^{(t+1)} &\leq \mathcal{A}_{\,i \,j k_2\cdots k_N} X_{ij}^{(t)} \,(\|X^{(t)}\|_{1,\infty}V)^{N-1} + (1-\mathcal{A}_{\,i \,j k_2\cdots k_N}) M_j^{(t)}(\|X^{(t)}\|_{1,\infty} V)^{N-1} \\&\leq^{\star} \mathcal{A}_{\,i \,j k_2\cdots k_N} X_{ij}^{(t)} + (1-\mathcal{A}_{\,i \,j k_2\cdots k_N}) M_j^{(t)}
\end{align*}
for each row $i$, there is $1 \geq \rho_i >0$, such that $X_{ij}^{(t)} = \rho_i \,m^{(t)}_j + (1-\rho_i)M^{(t)}_j$
\begin{align*}
    X_{ij}^{(t+1)}
    &\leq \varepsilon X_{ij}^{(t)} + (1-\varepsilon) M_j^{(t)} \\
    &\leq \varepsilon  (\rho_i \,m^{(t)}_j + (1-\rho_i)M^{(t)}_j) + (1-\varepsilon) M_j^{(t)} \\
    &=   \rho_i\,\varepsilon \,m^{(t)}_j + (1-\rho_i\,\varepsilon)M^{(t)}_j
\end{align*}
so $X_{ij}^{(t+1)} \leq \rho_i\,\varepsilon \,m^{(t)}_j + (1-\rho_i\,\varepsilon)M^{(t)}_j \qquad$ now let's do the case $t+2$:
\begin{align*}
    X_{ij}^{(t+2)} &\leq \mathcal{A'}_{\,i \,j k_2\cdots k_N} X_{ij}^{(t+1)} \,(\|X^{(t+1)}\|_{1,\infty}V)^{N-1} + (1-\mathcal{A'}_{\,i \,j k_2\cdots k_N}) M_j^{(t+1)}(\|X^{(t+1)}\|_{1,\infty} V)^{N-1} \\&\leq^{\star} \mathcal{A'}_{\,i \,j k_2\cdots k_N} X_{ij}^{(t+1)} + (1-\mathcal{A'}_{\,i \,j k_2\cdots k_N}) M_j^{(t)} \\
    &\leq \varepsilon\big(\rho_i\,\varepsilon \,m^{(t)}_j + (1-\rho_i\,\varepsilon)M^{(t)}_j \big) + (1-\varepsilon)M_j^{(t)} \\
    &\leq \rho_i\,\varepsilon^2 \,m^{(t)}_j + (1-\rho_i\,\varepsilon^2)M^{(t)}_j
\end{align*}
\noindent By iterating, we get $ \quad X^{(t+T)}_{i\,j} \leq \rho_i \,\varepsilon^T m^{(t)} + (1-\rho_i \,\varepsilon^T)M^{(t)}$
\qed


%% file: tex/annex5.tex
$\newline$ 
\section{Proof of \autoref{thm:lips} : Lipschitz upper bound for N-simplicial attention} \label{appx:lips}
\begin{tcolorbox}[enhanced, colback=red!10!white, colframe=red!80!black]

\lipthm*
\end{tcolorbox}
\noindent
For $x_1, \ldots, x_n$ the token representation (rows of features $X$), the $i^{th}$ row of the N-simplicial attention output is:
$$ f_i(X) = f_i(x_1, \ldots, x_n) = \underset{\mathcal{K} = (k_1\cdots k_N)}{\sum} \mathcal{A}_{i\mathcal{K}} \underset{m=1}{\overset{N}{\prod}}x_{k_m} W_V^{(m)}$$ \\
this is the row-wise version of the formula presented in \autoref{subsec:nsimp}, we will use below $K^{(m)}:=W_K^{(m)}$ \\
We want to extend the upper bound of \citealt{castin2024smoothattention} to higher order attention, we have :
\begin{align*}
    \text{Lip}(f_{|\mathcal{X}}) &= \underset{X\in\mathcal{X}}{\sup} \|D_Xf\|_2 \\
    \|D_Xf \|_2 &= \Big(\sum_{1\leq i \leq n} \|(D_Xf)(\varepsilon)_i \|_2^2 \Big)^{1/2} 
\end{align*}

\begin{align*}
    \|(D_Xf)(\varepsilon)_i \|_2 &= \| \underset{j=1}{\overset{n}{\sum}}  \pd{f_i}{x_j}(X) \,\varepsilon_j \|_2 \\
    &\leq  \underset{j=1}{\overset{n}{\sum}} \left\| \,\,
    \underset{\mathcal{K}}{\sum} \mathcal{A}_{i\mathcal{K}} \pd{}{x_j}(\underset{m=1}{\overset{N}{\prod}} x_{k_m} W_V^{(m)}) \,\varepsilon_j \,\,\,\, + \,\,\,\, \underset{\mathcal{K}}{\sum} \pd{\mathcal{A}_{i\mathcal{K}}}{x_j}(X) \,\,\bigg(\underset{m=1}{\overset{N}{\prod}}x_{k_m}W_V^{(m)}\bigg) \varepsilon_j  \right\|_2 \\
    &\leq  \underset{j=1}{\overset{n}{\sum}} \left( \,\,
    \bigg\|\underset{\mathcal{K}}{\sum}  \mathcal{A}_{i\mathcal{K}}  \pd{}{x_j}(\underset{m=1}{\overset{N}{\prod}} x_{k_m} W_V^{(m)}) \,\varepsilon_j \bigg\|_2 \,\,\,\, + \,\,\,\, \bigg\| \underset{\mathcal{K}}{\sum}  \pd{\mathcal{A}_{i\mathcal{K}}}{x_j}(X) \,\,\bigg(\underset{m=1}{\overset{N}{\prod}}x_{k_m}W_V^{(m)}\bigg) \varepsilon_j \bigg\|_2  \right) \\
    &\text{using \autoref{lem:prod} and \autoref{lem:derivA} we have} \\
    &\leq \underset{j=1}{\overset{n}{\sum}} \big( N\,V^N\,R^{N-1} \,+ \sqrt{d} \, N \, V^N K^{N+1} R^{2N} \big) \\
    &\leq nNV^N\,R^{N-1} \big(1 + \sqrt{d} \,(KR)^{N+1} \big)
\end{align*}
thus we can conclude, as $(a+b)^2 \leq 2(a^2 + b^2)$ and $\|D_Xf \|_2 = \Big(\sum_{1\leq i \leq n} \|(D_Xf)(\varepsilon)_i \|_2^2 \Big)^{1/2} $
\[
\boxed{
    \text{Lip}(f_{|B^n_R}) \leq n\sqrt{2n} \,N\,V^N\,R^{N-1}\big(1 + d\,N^2 (KR)^{2(N+1)}\big)^{1/2}
}\]
\qed

\begin{lemma} \label[lemma]{lem:prod} under the same assumptions as \autoref{thm:lips}, \\ i.e. $X \in B^n_R$, $ V = {\max}_{1 \leq m \leq N} \| W_V^{(m)} \|_2$, $ K = {\max}_{0 \leq m \leq N} \| K^{(m)} \|_2$ and $\sum_{\mathcal{K}} \mathcal{A}_{i\mathcal{K}} = 1$ we have :
$$
\left\| \underset{\mathcal{K}}{\sum} \mathcal{A}_{i\mathcal{K}}  \pd{}{x_j}(\underset{m=1}{\overset{N}{\prod}} x_{k_m} W_V^{(m)}) \,\varepsilon_j \right\|_2 \leq \, N \,V^N R^{N-1}
$$
\end{lemma} \noindent \textbf{Proof.} we have
\begin{align*}
\bigg\|\underset{\mathcal{K}}{\sum}  \mathcal{A}_{i\mathcal{K}}  \pd{}{x_j}(\underset{m=1}{\overset{N}{\prod}} x_{k_m} W_V^{(m)}) \,\varepsilon_j \bigg\|_2 &= \underset{\mathcal{K}}{\sum} \bigg\| \mathcal{A}_{i\mathcal{K}} \underset{a=1}{\overset{N}{\sum}} \delta_{k_a = j} W_V^{(a)}\Big( \underset{m\neq a}{\overset{N}{\prod}}\ x_{k_m}W_V^{(m)} \Big)
\, \varepsilon_j \bigg\|_2 \, ... \\
&\leq^{(1)} \underset{\mathcal{K}}{\sum}  \mathcal{A}_{i\mathcal{K}} N\, V^N \, R^{N-1}
\, \left\|\varepsilon_j \right\|_2 \\
&\leq^{(2)} N\, V^N \, R^{N-1}
\end{align*}
because $(2):\, \underset{\mathcal{K}}{\sum}  \mathcal{A}_{i\mathcal{K}} \left\|\varepsilon_j \right\|_2 \leq 1 $ as $ \underset{\mathcal{K}}{\sum}  \mathcal{A}_{i\mathcal{K}}= 1 $ and $\|\varepsilon\|_F^{2} \leq 1$ \\
and $(1): \, \Big\| \underset{a=1}{\overset{N}{\sum}} \delta_{k_a = j} W_V^{(a)}\Big( \underset{m\neq a}{\overset{N}{\prod}}\ x_{k_m}W_V^{(m)} \Big) \Big\|_2 \leq \underset{a=1}{\overset{N}{\sum}} V \,(VR)^{N-1} \leq N\,V^N\,R^{N-1}$ \\
\qed
$\newline \newline$
\begin{lemma} \label[lemma]{lem:derivA} under the same assumptions as \autoref{thm:lips}, \\ i.e. $X \in B^n_R$, $ V = \underset{1 \leq m \leq N}{\max} \| W_V^{(m)} \|_2$ and $ K = \underset{0 \leq m \leq N}{\max} \| K^{(m)} \|_2$ we have :
$$
 \left\| \underset{\mathcal{K}}{\sum} \pd{\mathcal{A}_{i\mathcal{K}}}{x_j}(X) \,\,\bigg(\underset{m=1}{\overset{N}{\prod}}x_{k_m}W_V^{(m)}\bigg) \varepsilon_j \right\|_2 \leq \, \sqrt{d} \, N \, V^N K^{N+1} R^{2N}
$$
\end{lemma} \noindent \textbf{Proof.} to simplify, we will use the notation $e^\mathcal{K} := \exp(\frac{1}{\sqrt{d}} \underset{a=1}{\overset{d}{\sum}} x_i K_{:,a}^{(0)} \, \cdot \, x_{k_1} K_{:,a}^{(1)} \cdots x_{k_N} K_{:,a}^{(N)})$, \\ and respectively : $e^M := \exp(\frac{1}{\sqrt{d}} \underset{a=1}{\overset{d}{\sum}} x_i K_{:,a}^{(0)} \, \cdot \, x_{\mathbf{m}_1} K_{:,a}^{(1)} \cdots x_{\mathbf{m}_N} K_{:,a}^{(N)})$ \\
\begin{align*}
    \pd{\mathcal{A}_{i\mathcal{K}}}{x_j}(X) &= \pd{}{x_j}\Big( \frac{e^\mathcal{K}}{\sum_M e^M}\Big) \\
    &= \frac{\pd{e^\mathcal{K}}{x_j} (\sum_M e^M)}{(\sum_M e^M)^2} - \frac{e^\mathcal{K} \pd{(\sum_M e^M)}{x_j} }{(\sum_M e^M)^2} \\
    &= \frac{1}{\sum_M e^M}\pd{(e^\mathcal{K})}{x_j} - \mathcal{A}_{i\mathcal{K}}\frac{1 }{\sum_M e^M} \sum_M \pd{(e^M)}{x_j}
\end{align*}

\noindent $\pd{(e^\mathcal{K})}{x_j} = \pd{}{x_j}\big(\exp(\frac{1}{\sqrt{d}} \underset{a=1}{\overset{d}{\sum}} x_i K_{:,a}^{(0)} \, \cdot \, x_{k_1} K_{:,a}^{(1)} \cdots x_{k_N} K_{:,a}^{(N)})\big)  = \pd{}{x_j}\big((\frac{1}{\sqrt{d}} \underset{a=1}{\overset{d}{\sum}} x_i K_{:,a}^{(0)} \, \cdot \, x_{k_1} K_{:,a}^{(1)} \cdots x_{k_N} K_{:,a}^{(N)})\big) e^\mathcal{K} \newline := \pd{B_\mathcal{K}}{x_j} e^\mathcal{K} \quad$ using the notation $B_\mathcal{K} =\frac{1}{\sqrt{d}} \underset{a=1}{\overset{d}{\sum}} x_i K_{:,a}^{(0)} \, \cdot \, x_{k_1} K_{:,a}^{(1)} \cdots x_{k_N} K_{:,a}^{(N)} $

\begin{align*}
    \pd{\mathcal{A}_{i\mathcal{K}}}{x_j}(X) &= \frac{1}{\sum_M e^M}\pd{(e^\mathcal{K})}{x_j} - \mathcal{A}_{i\mathcal{K}}\frac{1 }{\sum_M e^M} \sum_M \pd{(e^M)}{x_j} \\
    &= \frac{1}{\sum_M e^M}\pd{(B_\mathcal{K})}{x_j}e^\mathcal{K} - \mathcal{A}_{i\mathcal{K}}\frac{1}{\sum_M e^M} \sum_M \pd{(B_M)}{x_j} e^M  \\
    &= \mathcal{A}_{i\mathcal{K}} \pd{(B_\mathcal{K})}{x_j} - \mathcal{A}_{i\mathcal{K}} \sum_M \mathcal{A}_{iM}\pd{(B_M)}{x_j}
\end{align*}
Thus, we have :
\begin{align*}
   \left\| \underset{\mathcal{K}}{\sum}  \pd{\mathcal{A}_{i\mathcal{K}}}{x_j}(X) \,\,\bigg(\underset{m=1}{\overset{N}{\prod}}x_{k_m}W_V^{(m)}\bigg) \varepsilon_j \right\|_2 
    &\leq V^NR^N \left\| \underset{\mathcal{K}}{\sum}  \pd{\mathcal{A}_{i\mathcal{K}}}{x_j}(X) \,\varepsilon_j\right\|_2 \\
    &\leq V^NR^N \left\| \underset{\mathcal{K}}{\sum}  \mathcal{A}_{i\mathcal{K}} \Big(\pd{(B_\mathcal{K})}{x_j} - \sum_M \mathcal{A}_{iM}\pd{(B_M)}{x_j} \Big) \,\varepsilon_j\right\|_2 \\
    &\leq^{(1)} V^NR^N \Big(\underset{\mathcal{K}}{\sum} \mathcal{A}_{i\mathcal{K}} \Big\| \pd{(B_\mathcal{K})}{x_j} - \sum_M \mathcal{A}_{iM}\pd{(B_M)}{x_j} \Big\|_2^2\Big)^{1/2} \, \Big(\underset{\mathcal{K}}{\sum} \mathcal{A}_{i\mathcal{K}} \|\varepsilon_j\|^2_2 \Big)^{1/2} \\
    &\leq V^NR^N \Big(\underset{\mathcal{K}}{\sum} \mathcal{A}_{i\mathcal{K}} \Big\| \pd{(B_\mathcal{K})}{x_j} - \sum_M \mathcal{A}_{iM}\pd{(B_M)}{x_j} \Big\|_2^2\Big)^{1/2}
\end{align*}
$(1): $ via the inequality of Cauchy-Schwarz. \\
and $\underset{\mathcal{K}}{\sum} \mathcal{A}_{i\mathcal{K}} \Big\| \pd{(B_\mathcal{K})}{x_j} - \sum_M \mathcal{A}_{iM}\pd{(B_M)}{x_j} \Big\|_2^2 = \mathbb{E}_{\mathcal{A}_i}[ \|  \pd{(B_\mathcal{K})}{x_j} - \mathbb{E}_{\mathcal{A}_i}[\pd{(B_\mathcal{K})}{x_j}]\|^2_2] = \mathrm{Var}_{\mathcal{A}_i}\left(g_{\mathcal{K}}(x_1, \dots, x_n)\right)
\leq r^2$ 
\noindent for $r$ such that $\forall \mathcal{K}, \forall X\in B^n_R,  \pd{(B_\mathcal{K})}{x_j} = g_{\mathcal{K}}(x_1, \cdots, x_n) \leq r$, we have such bound verified for $r = \sqrt{d}`\,N\, K^{N+1} R^N$ so:
\begin{align*}
   \left\| \underset{\mathcal{K}}{\sum}  \pd{\mathcal{A}_{i\mathcal{K}}}{x_j}(X) \,\,\bigg(\underset{m=1}{\overset{N}{\prod}}x_{k_m}W_V^{(m)}\bigg) \varepsilon_j \right\|_2 
    &\leq V^NR^N r \\
    &\leq \sqrt{d}\, N \,V^N \, K^{N+1} R^{2N}
\end{align*}
\qed

%% file: tex/annex4.tex
\newpage
\section{Higher-order Message Passing and N-simplicial attention} \label{appx:hom}
\begin{figure*}[h]
\centering
\includegraphics[width=0.9\textwidth]{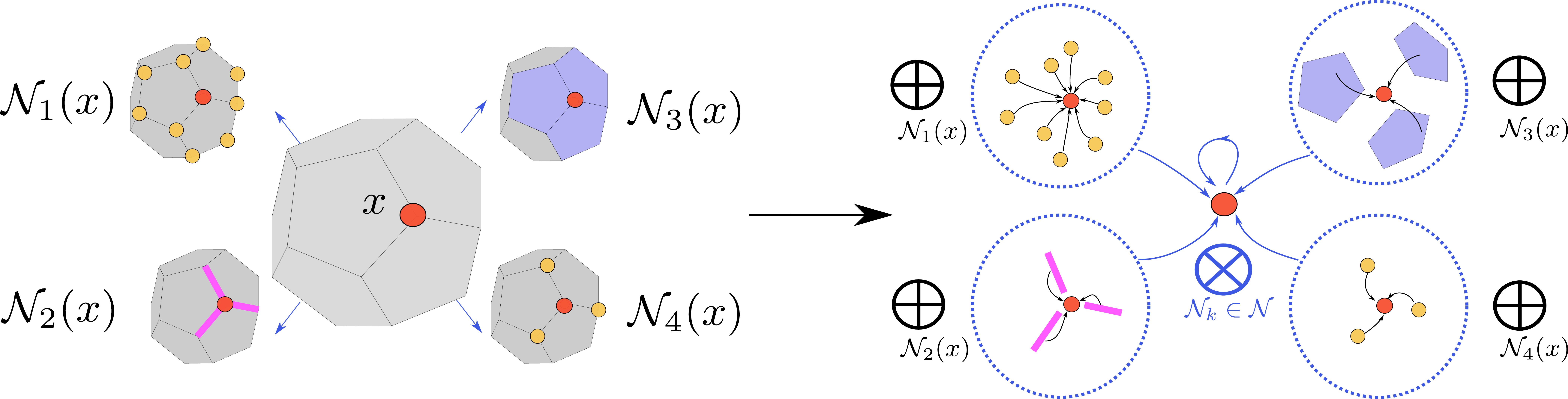}
\caption{an illustration of higher-order message passing from \citealt{hajij2023topologicaldeeplearninggoing}, p.34 \\
on the left, neighborhoods of different orders, on the right, the message-passing inside those neighborhoods.}
\label{fig:homp}
\end{figure*}

An important idea throughout this paper is the close link between attention in transformer and message-passing in Graph Neural Network (cf. \citealt{joshi2025transformersgraphneuralnetworks}). Following this paradigm, in this section, we propose to try to build a bridge between N-simplicial attention (\S \ref{subsec:nsimp}) and previous works in Higher-order Message Passing (HOMP) (\citealt{hajij2023topologicaldeeplearninggoing}, \citealt{goh2022simplicialattentionnetworks}, \citealt{morris2021weisfeilerlemanneuralhigherorder}), and to show more clearly the differences. \\ Let's first make clear the definition of usual message-passing:
\begin{tcolorbox} \begin{definition*}[Message Passing]\label{def:mp}
Each node $i\in \mathcal{V}(\mathcal{G})$ iteratively updates its representation $h_i^{(t)} = X^{(t)}_{i,:}$ to the next layer $h_i^{(t+1)}$ by aggregating messages from the representation of all of its local neighbors $j \in \mathcal{N}_i(\mathcal{G})$. \\
First we \underline{construct the message} from $j\to i$ : $$m_{ij}^{(t)} = \psi(h^{(t)}_i, h^{(t)}_j) \qquad \forall j \in  \mathcal{N}_i \qquad \text{with } \psi : \mathbb{R}^{2d} \to \mathbb{R}^d$$ 
Then we \underline{aggregate} across all the $j\in\mathcal{N}_i(\mathcal{G})$ neighbors of $i$ : $$ m^{(t)}_i  = \bigoplus_{j \in \mathcal{N}_i} m_{ij}^{(t)} \qquad \quad \,\,\,\quad (\bigoplus \text{ permutation-invariant})$$ 
Finally we \underline{update} the node representation:  $h^{(t+1)}_{i} = \phi(h^{(t)}_{i}, m^{(t)}_i)$ \hfill
(\citealt{battaglia2018relationalinductivebiasesdeep}, \citealt{joshi2025transformersgraphneuralnetworks})
\end{definition*} \end{tcolorbox}
\begin{remarks*}
As \cite{joshi2025transformersgraphneuralnetworks} mentions, this framework describes different models, see below the main examples: \vspace{-0.2cm} \begin{enumerate}
        \item GNN : $\psi$ is an MLP, $\bigoplus$ a sum, and $\phi$ is another MLP, $\mathcal{G}$ the given graph structure
        \item GAT : $\psi$ is a local-attention function (see below), $\bigoplus$ a sum, $\phi$ a sum, $\mathcal{G}$ idem (the given graph).
        $$
        \psi(h^{(t)}_i, h^{(t)}_j) = \frac{\exp(W_Q^{(t)} h^{(t)}_i \cdot W_K^{(t)} h^{(t)}_j)}{ \sum_{j' \in \mathcal{N}_i}\exp(W_Q^{(t)} h^{(t)}_i \cdot W_K^{(t)} h^{(t)}_{j'})} W_V^{(t)} h^{(t)}_j
        $$
        note that due to the normalization, the $\psi(h^{(t)}_i, h^{(t)}_j)$ formulation is not completely rigorous, but as it's the same for each message of the neighborhood, we could put it in $\bigoplus$ anyway, (not done to simplify the notation). \\
        \item Attention :  $\mathcal{G}$ is the complete graph of the tokens, $\psi$ is a global-attention function (see below)
        $$
        \psi(h^{(t)}_i, h^{(t)}_j) = \frac{\exp(W_Q^{(t)} h^{(t)}_i \cdot W_K^{(t)} h^{(t)}_j)}{ \sum_{j' \in \mathcal{V}(\mathcal{G})}\exp(W_Q^{(t)} h^{(t)}_i \cdot W_K^{(t)} h^{(t)}_{j'})} W_V^{(t)} h^{(t)}_j
        $$
         $\bigoplus$ is a sum : $m_i^{(t)} = \sum_{j \in \mathcal{V}(\mathcal{G})} \psi(h^{(t)}_i,h^{(t)}_j) \qquad$ and in a Transformer, $\phi$ is such that :
         $$
         \phi(h^{(t)}_i, m_i^{(t)}) = \text{MLP}(\text{LayerNorm}(h^{(t)}_i + m_i^{(t)}))
         $$
         for masked attention, $\mathcal{G}$ is now the graph of adjacency following the mask.
    \end{enumerate}
\end{remarks*}
\noindent \underline{For N-simplicial attention}:\\
$\mathcal{G}$ is the complete N-uniform hypergraph or the mask, $\psi$ replaces the dot-product by its N-linear generalization :
        $$
        \psi(h^{(t)}_i, h^{(t)}_{j_1}, \cdots, h^{(t)}_{j_N}) = \frac{\exp(W_Q^{(t)} h^{(t)}_i \cdot W_{K_1}^{(t)} h^{(t)}_{j_1} \cdots W_{K_N}^{(t)} h^{(t)}_{j_N})}{ \underset{(k_1, \cdots, k_N) \in \mathcal{V}(\mathcal{G})^N}{\sum}\exp(W_Q^{(t)} h^{(t)}_{i} \cdot W_{K_1}^{(t)} h^{(t)}_{k_1} \cdots W_{K_N}^{(t)} h^{(t)}_{k_N})} (W_{V_1}^{(t)} h^{(t)}_{j_1} \cdots W_{V_N}^{(t)} h^{(t)}_{j_N})
        $$
$\bigoplus$ is the sum of the neighboring simplexes' messages : $m_i^{(t)} = {\sum}_{(j_1, \cdots, j_N) \in \mathcal{V}(\mathcal{G})^N} \,\, \psi(h^{(t)}_i,h^{(t)}_{j_1}, \cdots, h^{(t)}_{j_N})$ \\
$\phi$ is usually the same as for attention, $\phi(h^{(t)}_i, m_i^{(t)}) = \text{MLP}(\text{LayerNorm}(h^{(t)}_i + m_i^{(t)}))$ \\ 
The only difference making message-passing not able to describe N-simplicial attention is that we construct messages $\psi$ from simplexes of nodes altogether and not just pairs. Let's try to link it to higher-order MP:
\begin{tcolorbox}
\begin{definition*}[Higher-order Message Passing]  \hfill (from \citealt{hajij2023topologicaldeeplearninggoing}) \\
    \noindent Let $\mathcal{C}$ be a Cell-Complex. Let $ \mathcal{N} = \{ \mathcal{N}_1, \dots, \mathcal{N}_n \} $ be a set of neighborhood functions defined on $\mathcal{C}$. \\
    Let $x$ be a cell and a neighbor $y \in \mathcal{N}_k(x)$ for some $\mathcal{N}_k \in \mathcal{N}$. \\
    Denote by $ \mathcal{N}(x) $ the multi-set $\{ \{ \mathcal{N}_1(x), \dots, \mathcal{N}_n(x) \}\} $, and by $h^{(t)}_x$ the representation of $x$ at layer $t$. \\
    Higher-order message passing on $\mathcal{C}$, induced by $\mathcal{N}$, is defined via the following four update rules:
    \begin{align*}
        m_{x,y} &= \psi_{\mathcal{N}_k}(h^{(t)}_x, h^{(t)}_y) \\
        m_x^k &= \bigoplus_{y \in \mathcal{N}_k(x)} m_{x,y} \qquad \qquad (\bigoplus \text{ permutation-invariant})  \\
        m_x &= \bigotimes_{\mathcal{N}_k \in \mathcal{N}} m_x^k \\
        h^{(t+1)}_x &= \phi(h^{(t)}_x, m_x)
    \end{align*}
\noindent where $ \bigotimes$ is an aggregation function over all the neighborhoods of different order of $x$ \hfill see \autoref{fig:homp}
\end{definition*}
\end{tcolorbox}

\noindent This is different than N-simplicial attention, as this keeps in memory the features of higher-order objects (cells), and do pairwise message passing among all of those higher-order objects. Instead of recreating them from the nodes features only (see \citealt{ballester2024attendingtopologicalspacescellular}, \citealt{zhou2024theoreticalexpressivepowerdesign} for pairwise attention between given higher-order objects). To do N-simplicial attention from Higher-order message passing, we would have to add additional rules on $h^{(t)}_x$ such that it correspond to what is obtained in N-simplicial attention, making N-simplicial attention new and different.

\section{Additional illustrative figures}
\begin{figure*}[h]
\centering
\includegraphics[width=0.7\textwidth]{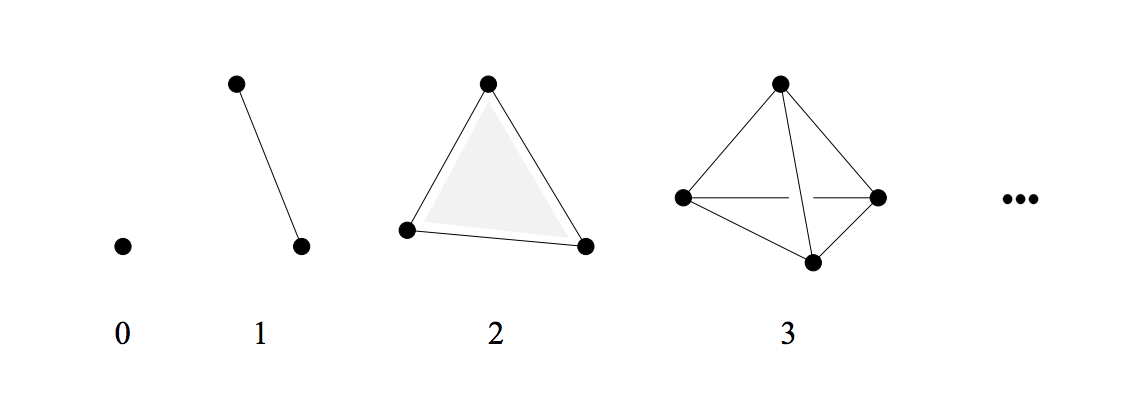}
\caption{a 0-simplex is a node/vertex, a 1-simplex is an edge, a 2-simplex is a triangle, a 3-simplex is a tetrahedron, etc. \\
modified from \textit{Geometry and Topology of Grid Generation}, H.Edelsbrunner, Spring 1999. Simplicial Complexes}
\label{fig:placeholder}
\end{figure*}

\begin{figure*}[h]
\centering
\includegraphics[width=0.75\textwidth]{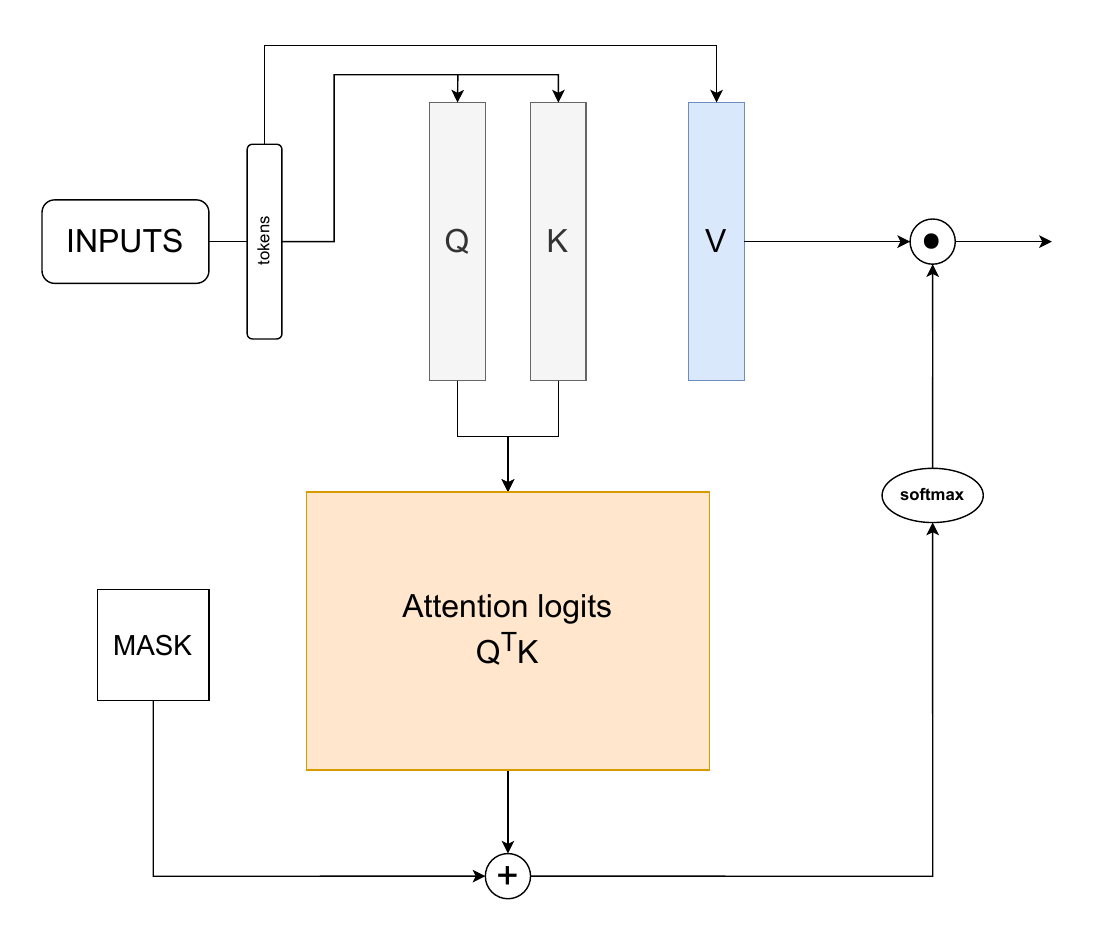}
\caption{An illustration of masked dot product attention.}
\label{fig:attn}
\end{figure*}
\begin{figure*}[h]
\centering
\includegraphics[width=0.75\textwidth]{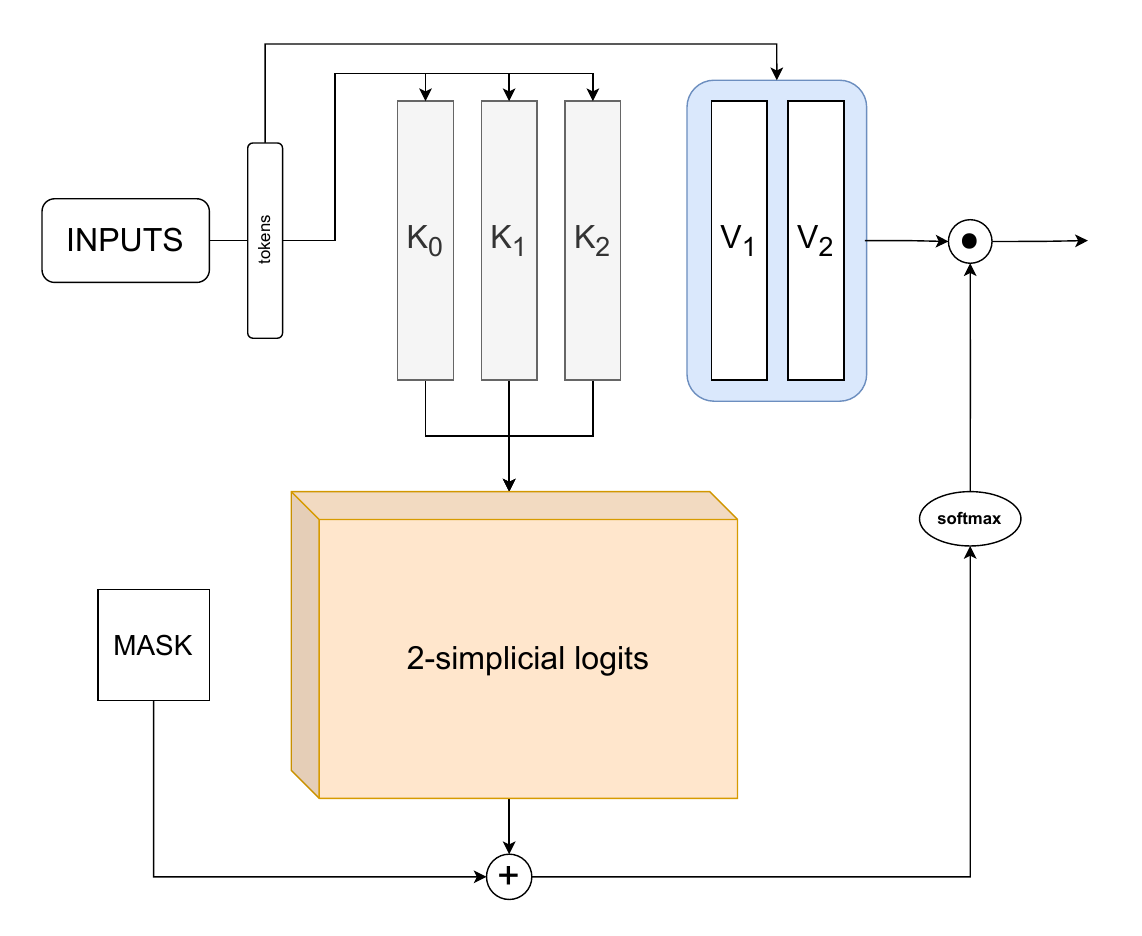}
\caption{An illustration of masked 2-simplicial attention.}
\label{fig:2simp}
\end{figure*}
\clearpage

\newpage
\section{Complementary remarks} \label{subsec:reduce}

\subsection{N-simplicial attention can simulate every lower order m-simplicial attention $(m\leq N)$}
N-simplicial attention can be reduced to the lower order simplicial attentions, indeed we can do $(k\leq N)$-simplicial attention, all at once with little changes: 
Let's note $\mathcal{A}^{(2)}$ the attention matrix, $\mathcal{A}^{(k)}$ the k-simplicial attention logits tensor, by adding a new row to $K^{(2)}$ fully composed of 1, we get a $n\times n\times(n+1)$ tensor with both the attention and the 2-simplicial attention logits.
\begin{equation*}
    \mathcal{A}^{(3)}_{ijk}  = \overset{d}{\underset{a=1}{\sum}}Q_{ia}K^{(1)}_{ja}K^{(2)}_{ka} \quad 
    \underset{\mathrm{if} \,\, \,K^{(2)}_{ka}=1}{=} \quad\overset{d}{\underset{a=1}{\sum}}Q_{ia}K^{(1)}_{ja}  \times1 \,\,\,\,\, 
    = \,A^{(2)}_{ij}
\end{equation*}
\noindent
In the same way, we can do that for each of $K^{(i)}\, (2\leq i \leq N)$ computing all the simplicial-attention under N at the same time. Adapting the dimensions of $V_{(i)}$. We get a logits tensor of dimension $n^2 + \cdots + n^{N+1}  \sim O(n^{N+1}) $. \\
For $N\leq M$, N-simplicial attention is a particular case of M-simplicial attention.

$\newline$
\subsection{Curvature of the line-graph}
\textbf{Why is it better to work on the edges than on the nodes ?} \citealt{Buterez_2025} recently published state-of-the-art results for graph-level regression tasks across wide range of datasets, by working directly on the edges features $[n_1||n_2||e_{12}]$ (the concatenation of the source node, the target node and their specific edge features) instead of the nodes, with a decoder-only transformer using the edge connectivity as the attention mask. We propose in this paragraph results that could explain why working on the line-graph (on the edges and their connectivity) improved the model predictions. 

\noindent Over-squashing \citep{alon2021bottleneckgraphneuralnetworks} can be partly induced by the "bottleneck" tendency of a graph, itself correlated to the Forman-Ricci curvature of its edges \citep{topping2022understandingoversquashingbottlenecksgraphs}
\begin{figure}[H]
    \centering
    \begin{minipage}[t]{0.48\textwidth}
        \centering
            \centering
            \vspace{-4cm}
            \scalebox{0.55}{\input{img/linegraph}}
            \scalebox{0.44}{\input{img/linegraph2}}
        \captionsetup{justification=centering}
        \caption{Intuition, less bottlenecks in line-graphs}
    \end{minipage}
    \hfill
    \begin{minipage}[t]{0.48\textwidth}
        \hspace{0.2cm}
        \includegraphics[width=1.0\textwidth]{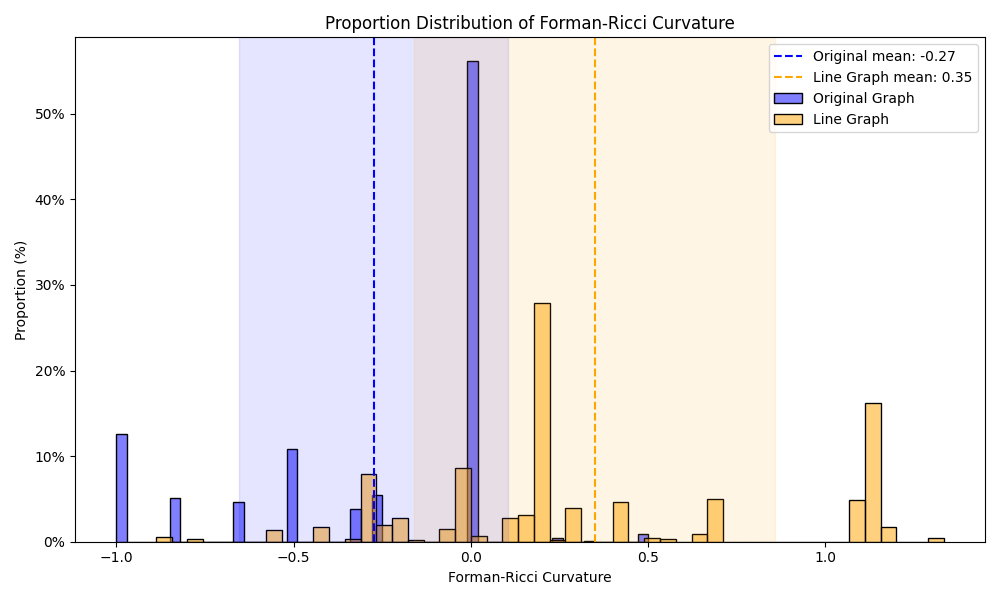}
        
        \captionsetup{justification=centering}
        \caption{Ricci curvature (graphs $\&$ line graphs) \\ QM9 dataset, $\geq$ 100 000 samples}
    \end{minipage}
    \label{fig:ricci_qm9}
\end{figure} \noindent We observe empirically, on the QM9 dataset (tested for more than 100 000 molecules) that the average Forman-Ricci curvature is increased for the line-graph in 100\% of the case. And the smallest Forman–Ricci curvature among the graph’s edges is lower than the smallest Ricci curvature among the edges of its line-graph in at least 95\% of the case (on the QM9 dataset). Both reducing the over-squashing tendency when working on the line-graphs. 
This is feasible as the molecule graph edge density is quite low (average edge density of 0.12 in the QM9 dataset), meaning that taking the line-graph doesn't increase the tokens count too much. 

%% file: img/linegraph.tex
\usetikzlibrary{shapes,fit,positioning,backgrounds,arrows.meta}

\tikzstyle{vertex} = [fill,shape=circle,node distance=80pt]
\tikzstyle{edge} = [fill,opacity=.5,fill opacity=.5,line cap=round, line join=round, line width=50pt]

\pgfdeclarelayer{background}
\pgfsetlayers{background,main}

\begin{tikzpicture}

\node[vertex] (u1) at (0,1.5) {};
\node[vertex] (u2) at (0,-1.5) {};
\node[vertex] (u3) at (2,0) {};
\node[vertex] (u4) at (4,0) {};
\node[vertex] (u5) at (6,1.5) {};
\node[vertex] (u6) at (6,-1.5) {};

\begin{scope}[edge/.style={draw, line width=2pt}]
\begin{pgfonlayer}{background}
\draw[edge, opacity=0.5] (u1) -- (u3) node[midway, above] {$\,e_1$};
\draw[edge, opacity=0.5] (u2) -- (u3) node[midway, below] {$\,e_2$};
\draw[edge, opacity=0.5] (u3) -- (u4) node[midway, below] {$e_3$};
\draw[edge, opacity=0.5] (u4) -- (u5) node[midway, above] {$e_4\,$};
\draw[edge, opacity=0.5] (u4) -- (u6) node[midway, below] {$e_5\,$};
\end{pgfonlayer}
\end{scope}

\begin{scope}[xshift=9cm]
\tikzstyle{vertex} = [fill,shape=rectangle,node distance=80pt]
\node[vertex, label=below:$e_1$ ] (u1) at (0,1.5) {};
\node[vertex, label=above:$e_2$ ] (u2) at (0,-1.5) {};
\node[vertex, label=above:$e_3$] (u3) at (2,0) {};
\node[vertex, label=below: $e_4$] (u5) at (4,1.5) {};
\node[vertex, label=above: $e_5$] (u6) at (4,-1.5) {};

\begin{scope}[edge/.style={draw, line width=2pt}]
\begin{pgfonlayer}{background}
\draw[edge, opacity=0.5] (u1) -- (u3);
\draw[edge, opacity=0.5] (u2) -- (u3);
\draw[edge, opacity=0.5] (u3) -- (u5);
\draw[edge, opacity=0.5] (u3) -- (u6);
\end{pgfonlayer}
\end{scope}
\end{scope}

\draw[->, thick] (6.5,0) -- (8.5,0) node[midway, above]{line graph};

\end{tikzpicture}

%% file: img/linegraph2.tex
\usetikzlibrary{shapes,fit,positioning,backgrounds,arrows.meta}

\tikzstyle{vertex} = [fill,shape=circle,node distance=80pt]
\tikzstyle{edge} = [fill,opacity=.5,fill opacity=.5,line cap=round, line join=round, line width=50pt]

\pgfdeclarelayer{background}
\pgfsetlayers{background,main}

\begin{tikzpicture}

\node[vertex] (u1) at (0,1.5) {};
\node[vertex] (u2) at (0,-1.5) {};
\node[vertex] (u3) at (2,0) {};
\node[vertex] (u4) at (4,0.5) {};
\node[vertex] (u7) at (6,0) {};
\node[vertex] (u5) at (8,1.5) {};
\node[vertex] (u6) at (8,-1.5) {};

\begin{scope}[edge/.style={draw, line width=2pt}]
\begin{pgfonlayer}{background}
\draw[edge, opacity=0.5] (u1) -- (u3) node[midway, above] {$\,e_1$};
\draw[edge, opacity=0.5] (u2) -- (u3) node[midway, below] {$\,e_2$};
\draw[edge, opacity=0.5] (u3) -- (u4) node[midway, below] {$e_3$};
\draw[edge, opacity=0.5] (u4) -- (u7) node[midway, below] {$e_4$};
\draw[edge, opacity=0.5] (u7) -- (u5) node[midway, above] {$e_5\,$};
\draw[edge, opacity=0.5] (u7) -- (u6) node[midway, below] {$e_6\,$};
\end{pgfonlayer}
\end{scope}

\begin{scope}[xshift=11cm]
\tikzstyle{vertex} = [fill,shape=rectangle,node distance=80pt]
\node[vertex, label=below:$e_1$ ] (u1) at (0.42,1.5) {};
\node[vertex, label=above:$e_2$ ] (u2) at (0.42,-1.5) {};
\node[vertex, label=above:$e_3$] (u3) at (2,0) {};
\node[vertex, label=above:$e_4$] (u4) at (4,0) {};
\node[vertex, label=below: $e_5$] (u5) at (6,1.5) {};
\node[vertex, label=above: $e_6$] (u6) at (6,-1.5) {};

\begin{scope}[edge/.style={draw, line width=2pt}]
\begin{pgfonlayer}{background}
\draw[edge, opacity=0.5] (u1) -- (u3);
\draw[edge, opacity=0.5] (u2) -- (u3);
\draw[edge, opacity=0.5] (u3) -- (u4);
\draw[edge, opacity=0.5] (u4) -- (u5);
\draw[edge, opacity=0.5] (u4) -- (u6);
\end{pgfonlayer}
\end{scope}
\end{scope}

\draw[->, thick] (8.5,0) -- (10.75,0) node[midway, above]{line graph};

\end{tikzpicture}

%% file: paper.bib
@misc{roy2025fastsimplex2simplicialattention,
      title={Fast and Simplex: 2-Simplicial Attention in Triton}, 
      author={Aurko Roy and Timothy Chou and Sai Surya Duvvuri and Sijia Chen and Jiecao Yu and Xiaodong Wang and Manzil Zaheer and Rohan Anil},
      year={2025},
      eprint={2507.02754},
      archivePrefix={arXiv},
      primaryClass={cs.LG},
      url={https://arxiv.org/abs/2507.02754}, 
}

@misc{clift2019logic2simplicialtransformer,
      title={Logic and the $2$-Simplicial Transformer}, 
      author={James Clift and Dmitry Doryn and Daniel Murfet and James Wallbridge},
      year={2019},
      eprint={1909.00668},
      archivePrefix={arXiv},
      primaryClass={cs.LG},
      url={https://arxiv.org/abs/1909.00668}, 
}

@misc{pappone2025beyondattention,
    author = {Francesco Pappone},
    title = {Beyond Attention as Graph},
    year = {2025},
    month = {October},
    day = {09},
    institution = {Università La Sapienza di Roma -- PSTP Technoscience},
    url = {https://publish.obsidian.md/the-tensor-throne/The+Graph+Side+of+Attention/Beyond+Attention+as+a+Graph},
    note = {Blogpost}
}

@misc{vaswani2023attentionneed,
      title={Attention Is All You Need}, 
      author={Ashish Vaswani and Noam Shazeer and Niki Parmar and Jakob Uszkoreit and Llion Jones and Aidan N. Gomez and Lukasz Kaiser and Illia Polosukhin},
      year={2017},
      eprint={1706.03762},
      archivePrefix={arXiv},
      primaryClass={cs.CL},
      url={https://arxiv.org/abs/1706.03762}, 
}

@article{Buterez_2025,
   title={An end-to-end attention-based approach for learning on graphs},
   url={http://dx.doi.org/10.1038/s41467-025-60252-z},
   journal={Nature Communications},
   publisher={Springer Science and Business Media LLC},
   author={Buterez, David and Janet, Jon Paul and Oglic, Dino and Liò, Pietro},
   year={2025},
}

@misc{topping2022understandingoversquashingbottlenecksgraphs,
      title={Understanding over-squashing and bottlenecks on graphs via curvature}, 
      author={Jake Topping and Francesco Di Giovanni and Benjamin Paul Chamberlain and Xiaowen Dong and Michael M. Bronstein},
      year={2022},
      eprint={2111.14522},
      archivePrefix={arXiv},
      primaryClass={stat.ML},
      url={https://arxiv.org/abs/2111.14522}, 
}

@inproceedings{Giraldo_2023, 
    series={CIKM’23},
    title={On the Trade-off between Over-smoothing and Over-squashing in Deep Graph Neural Networks},
    url={http://dx.doi.org/10.1145/3583780.3614997},
    DOI={10.1145/3583780.3614997},
    booktitle={Proceedings of the 32nd ACM International Conference on Information and Knowledge Management},
    author={Giraldo, Jhony H. and Skianis, Konstantinos and Bouwmans, Thierry and Malliaros, Fragkiskos D.},
    year={2023},
    collection={CIKM’23} }

@misc{wu2024roleattentionmaskslayernorm,
      title={On the Role of Attention Masks and LayerNorm in Transformers}, 
      author={Xinyi Wu and Amir Ajorlou and Yifei Wang and Stefanie Jegelka and Ali Jadbabaie},
      year={2024},
      eprint={2405.18781},
      archivePrefix={arXiv},
      primaryClass={cs.LG},
      url={https://arxiv.org/abs/2405.18781}, 
}

@misc{gharan2012universalupperboundgraph,
      title={A Universal upper bound on Graph Diameter based on Laplacian Eigenvalues}, 
      author={Shayan Oveis Gharan and Luca Trevisan},
      year={2012},
      eprint={1212.2701},
      archivePrefix={arXiv},
      primaryClass={cs.DM},
      url={https://arxiv.org/abs/1212.2701}, 
}

@misc{dong2023attentionneedpureattention,
      title={Attention is Not All You Need: Pure Attention Loses Rank Doubly Exponentially with Depth}, 
      author={Yihe Dong and Jean-Baptiste Cordonnier and Andreas Loukas},
      year={2023},
      eprint={2103.03404},
      archivePrefix={arXiv},
      primaryClass={cs.LG},
      url={https://arxiv.org/abs/2103.03404}, 
}

@misc{bae2025mixtureofrecursionslearningdynamicrecursive,
      title={Mixture-of-Recursions: Learning Dynamic Recursive Depths for Adaptive Token-Level Computation}, 
      author={Sangmin Bae and Yujin Kim and Reza Bayat and Sungnyun Kim and Jiyoun Ha and Tal Schuster and Adam Fisch and Hrayr Harutyunyan and Ziwei Ji and Aaron Courville and Se-Young Yun},
      year={2025},
      eprint={2507.10524},
      archivePrefix={arXiv},
      primaryClass={cs.CL},
      url={https://arxiv.org/abs/2507.10524}, 
}

@misc{raposo2024mixtureofdepthsdynamicallyallocatingcompute,
      title={Mixture-of-Depths: Dynamically allocating compute in transformer-based language models}, 
      author={David Raposo and Sam Ritter and Blake Richards and Timothy Lillicrap and Peter Conway Humphreys and Adam Santoro},
      year={2024},
      eprint={2404.02258},
      archivePrefix={arXiv},
      primaryClass={cs.LG},
      url={https://arxiv.org/abs/2404.02258}, 
}

@misc{alon2021bottleneckgraphneuralnetworks,
      title={On the Bottleneck of Graph Neural Networks and its Practical Implications}, 
      author={Uri Alon and Eran Yahav},
      year={2021},
      eprint={2006.05205},
      archivePrefix={arXiv},
      primaryClass={cs.LG},
      url={https://arxiv.org/abs/2006.05205}, 
}

@misc{wang2024auxiliarylossfreeloadbalancingstrategy,
      title={Auxiliary-Loss-Free Load Balancing Strategy for Mixture-of-Experts}, 
      author={Lean Wang and Huazuo Gao and Chenggang Zhao and Xu Sun and Damai Dai},
      year={2024},
      eprint={2408.15664},
      archivePrefix={arXiv},
      primaryClass={cs.LG},
      url={https://arxiv.org/abs/2408.15664}, 
}

@misc{zhou2022mixtureofexpertsexpertchoicerouting,
      title={Mixture-of-Experts with Expert Choice Routing}, 
      author={Yanqi Zhou and Tao Lei and Hanxiao Liu and Nan Du and Yanping Huang and Vincent Zhao and Andrew Dai and Zhifeng Chen and Quoc Le and James Laudon},
      year={2022},
      eprint={2202.09368},
      archivePrefix={arXiv},
      primaryClass={cs.LG},
      url={https://arxiv.org/abs/2202.09368}, 
}

@misc{deepseekai2024deepseekv32,
      title={DeepSeek-V3.2-Exp: Boosting Long-Context Efficiency with DeepSeek Sparse Attention}, 
      author={DeepSeek-AI},
      year={2025},
      url={https://github.com/deepseek-ai/DeepSeek-V3.2-Exp}
}

@misc{joshi2025transformersgraphneuralnetworks,
      title={Transformers are Graph Neural Networks}, 
      author={Chaitanya K. Joshi},
      year={2025},
      eprint={2506.22084},
      archivePrefix={arXiv},
      primaryClass={cs.LG},
      url={https://arxiv.org/abs/2506.22084}, 
}

@misc{shao2023unifyingoversmoothingoversquashinggraph,
      title={Unifying over-smoothing and over-squashing in graph neural networks: A physics informed approach and beyond}, 
      author={Zhiqi Shao and Dai Shi and Andi Han and Yi Guo and Qibin Zhao and Junbin Gao},
      year={2023},
      eprint={2309.02769},
      archivePrefix={arXiv},
      primaryClass={cs.LG},
      url={https://arxiv.org/abs/2309.02769}, 
}

@misc{rusch2023surveyoversmoothinggraphneural,
      title={A Survey on Oversmoothing in Graph Neural Networks}, 
      author={T. Konstantin Rusch and Michael M. Bronstein and Siddhartha Mishra},
      year={2023},
      eprint={2303.10993},
      archivePrefix={arXiv},
      primaryClass={cs.LG},
      url={https://arxiv.org/abs/2303.10993}, 
}

@misc{wu2024demystifyingoversmoothingattentionbasedgraph,
      title={Demystifying Oversmoothing in Attention-Based Graph Neural Networks}, 
      author={Xinyi Wu and Amir Ajorlou and Zihui Wu and Ali Jadbabaie},
      year={2024},
      eprint={2305.16102},
      archivePrefix={arXiv},
      primaryClass={cs.LG},
      url={https://arxiv.org/abs/2305.16102}, 
}

@misc{barbero2025llmsattendtoken,
      title={Why do LLMs attend to the first token?}, 
      author={Federico Barbero and Álvaro Arroyo and Xiangming Gu and Christos Perivolaropoulos and Michael Bronstein and Petar Veličković and Razvan Pascanu},
      year={2025},
      eprint={2504.02732},
      archivePrefix={arXiv},
      primaryClass={cs.CL},
      url={https://arxiv.org/abs/2504.02732}, 
}

@misc{noci2022signalpropagationtransformerstheoretical,
      title={Signal Propagation in Transformers: Theoretical Perspectives and the Role of Rank Collapse}, 
      author={Lorenzo Noci and Sotiris Anagnostidis and Luca Biggio and Antonio Orvieto and Sidak Pal Singh and Aurelien Lucchi},
      year={2022},
      eprint={2206.03126},
      archivePrefix={arXiv},
      primaryClass={cs.LG},
      url={https://arxiv.org/abs/2206.03126}, 
}

@misc{keriven2022littlemuchtheoreticalanalysis,
      title={Not too little, not too much: a theoretical analysis of graph (over)smoothing}, 
      author={Nicolas Keriven},
      year={2022},
      eprint={2205.12156},
      archivePrefix={arXiv},
      primaryClass={stat.ML},
      url={https://arxiv.org/abs/2205.12156}, 
}

@misc{roth2024rankcollapsecausesoversmoothing,
      title={Rank Collapse Causes Over-Smoothing and Over-Correlation in Graph Neural Networks}, 
      author={Andreas Roth and Thomas Liebig},
      year={2024},
      eprint={2308.16800},
      archivePrefix={arXiv},
      primaryClass={cs.LG},
      url={https://arxiv.org/abs/2308.16800}, 
}

@misc{digiovanni2023oversquashingmessagepassingneural,
      title={On Over-Squashing in Message Passing Neural Networks: The Impact of Width, Depth, and Topology}, 
      author={Francesco Di Giovanni and Lorenzo Giusti and Federico Barbero and Giulia Luise and Pietro Lio' and Michael Bronstein},
      year={2023},
      eprint={2302.02941},
      archivePrefix={arXiv},
      primaryClass={cs.LG},
      url={https://arxiv.org/abs/2302.02941}, 
}

@misc{su2023roformerenhancedtransformerrotary,
      title={RoFormer: Enhanced Transformer with Rotary Position Embedding}, 
      author={Jianlin Su and Yu Lu and Shengfeng Pan and Ahmed Murtadha and Bo Wen and Yunfeng Liu},
      year={2023},
      eprint={2104.09864},
      archivePrefix={arXiv},
      primaryClass={cs.CL},
      url={https://arxiv.org/abs/2104.09864}, 
}

@misc{kazemnejad2023impactpositionalencodinglength,
      title={The Impact of Positional Encoding on Length Generalization in Transformers}, 
      author={Amirhossein Kazemnejad and Inkit Padhi and Karthikeyan Natesan Ramamurthy and Payel Das and Siva Reddy},
      year={2023},
      eprint={2305.19466},
      archivePrefix={arXiv},
      primaryClass={cs.CL},
      url={https://arxiv.org/abs/2305.19466}, 
}

@misc{barbero2025roundroundgomakes,
      title={Round and Round We Go! What makes Rotary Positional Encodings useful?}, 
      author={Federico Barbero and Alex Vitvitskyi and Christos Perivolaropoulos and Razvan Pascanu and Petar Veličković},
      year={2025},
      eprint={2410.06205},
      archivePrefix={arXiv},
      primaryClass={cs.CL},
      url={https://arxiv.org/abs/2410.06205}, 
}

@misc{touvron2023llamaopenefficientfoundation,
      title={LLaMA: Open and Efficient Foundation Language Models}, 
      author={LLaMA and Hugo Touvron and Thibaut Lavril and Gautier Izacard and Xavier Martinet and Marie-Anne Lachaux and Timothée Lacroix and Baptiste Rozière and Naman Goyal and Eric Hambro and Faisal Azhar and Aurelien Rodriguez and Armand Joulin and Edouard Grave and Guillaume Lample},
      year={2023},
      eprint={2302.13971},
      archivePrefix={arXiv},
      primaryClass={cs.CL},
      url={https://arxiv.org/abs/2302.13971}, 
}

@misc{deepseekai2025deepseekv3technicalreport,
      title={DeepSeek-V3 Technical Report}, 
      author={DeepSeek-AI},
        year={2025},
      eprint={2412.19437},
      archivePrefix={arXiv},
      primaryClass={cs.CL},
      url={https://arxiv.org/abs/2412.19437}, 
}

@misc{gemmateam2024gemmaopenmodelsbased,
      title={Gemma: Open Models Based on Gemini Research and Technology}, 
      author={Gemma},
      year={2024},
      eprint={2403.08295},
      archivePrefix={arXiv},
      primaryClass={cs.CL},
      url={https://arxiv.org/abs/2403.08295}, 
}

@misc{wang2024lengthgeneralizationcausaltransformers,
      title={Length Generalization of Causal Transformers without Position Encoding}, 
      author={Jie Wang and Tao Ji and Yuanbin Wu and Hang Yan and Tao Gui and Qi Zhang and Xuanjing Huang and Xiaoling Wang},
      year={2024},
      eprint={2404.12224},
      archivePrefix={arXiv},
      primaryClass={cs.CL},
      url={https://arxiv.org/abs/2404.12224}, 
}

@misc{sanford2023representationalstrengthslimitationstransformers,
      title={Representational Strengths and Limitations of Transformers}, 
      author={Clayton Sanford and Daniel Hsu and Matus Telgarsky},
      year={2023},
      eprint={2306.02896},
      archivePrefix={arXiv},
      primaryClass={cs.LG},
      url={https://arxiv.org/abs/2306.02896}, 
}

@misc{geshkovski2024emergenceclustersselfattentiondynamics,
      title={The emergence of clusters in self-attention dynamics}, 
      author={Borjan Geshkovski and Cyril Letrouit and Yury Polyanskiy and Philippe Rigollet},
      year={2024},
      eprint={2305.05465},
      archivePrefix={arXiv},
      primaryClass={cs.LG},
      url={https://arxiv.org/abs/2305.05465}, 
}

@misc{cimetière2025localmaxdynamicsattentiontransformers,
      title={Localmax dynamics for attention in transformers and its asymptotic behavior}, 
      author={Henri Cimetière and Maria Teresa Chiri and Bahman Gharesifard},
      year={2025},
      eprint={2509.15958},
      archivePrefix={arXiv},
      primaryClass={cs.CL},
      url={https://arxiv.org/abs/2509.15958}, 
}

@misc{karagodin2024clusteringcausalattentionmasking,
      title={Clustering in Causal Attention Masking}, 
      author={Nikita Karagodin and Yury Polyanskiy and Philippe Rigollet},
      year={2024},
      eprint={2411.04990},
      archivePrefix={arXiv},
      primaryClass={cs.LG},
      url={https://arxiv.org/abs/2411.04990}, 
}

@misc{elbayad2020depthadaptivetransformer,
      title={Depth-Adaptive Transformer}, 
      author={Maha Elbayad and Jiatao Gu and Edouard Grave and Michael Auli},
      year={2020},
      eprint={1910.10073},
      archivePrefix={arXiv},
      primaryClass={cs.CL},
      url={https://arxiv.org/abs/1910.10073}, 
}

@misc{schuster2022confidentadaptivelanguagemodeling,
      title={Confident Adaptive Language Modeling}, 
      author={Tal Schuster and Adam Fisch and Jai Gupta and Mostafa Dehghani and Dara Bahri and Vinh Q. Tran and Yi Tay and Donald Metzler},
      year={2022},
      eprint={2207.07061},
      archivePrefix={arXiv},
      primaryClass={cs.CL},
      url={https://arxiv.org/abs/2207.07061}, 
}

@inproceedings{Elhoushi_2024,
   title={LayerSkip: Enabling Early Exit Inference and Self-Speculative Decoding},
   url={http://dx.doi.org/10.18653/v1/2024.acl-long.681},
   booktitle={Proceedings of the 62nd Annual Meeting of the Association for Computational Linguistics},
   author={Elhoushi, Mostafa and Shrivastava, Akshat and Liskovich, Diana and Hosmer, Basil and Wasti, Bram and Lai, Liangzhen and Mahmoud, Anas and Acun, Bilge and Agarwal, Saurabh and Roman, Ahmed and Aly, Ahmed and Chen, Beidi and Wu, Carole-Jean},
   year={2024}
}

@inproceedings{10.5555/2987189.2987330,
    author = {Sperduti, Alessandro},
    title = {Encoding labeled graphs by labeling RAAM},
    year = {1993},
    series = {NIPS'93},
    url = {https://dl.acm.org/doi/10.5555/2987189.2987330}
}

@ARTICLE{4700287,
  author={Scarselli, Franco and Gori, Marco and Tsoi, Ah Chung and Hagenbuchner, Markus and Monfardini, Gabriele},
  journal={IEEE Transactions on Neural Networks}, 
  title={The Graph Neural Network Model}, 
  year={2009},
  doi={10.1109/TNN.2008.2005605},
  url={https://ieeexplore.ieee.org/document/4700287},
}

@article{Gori2005ANM,
  title={A new model for learning in graph domains},
  author={Marcelo Gori and Gabriele Monfardini and F. Scarselli},
  journal={Proceedings. 2005 IEEE International Joint Conference on Neural Networks, 2005.},
  year={2005},
  url={https://api.semanticscholar.org/CorpusID:275758328}
}

@misc{abella2025asymptoticbehaviorattentiontransformers,
      title={The Asymptotic Behavior of Attention in Transformers}, 
      author={Rodríguez Abella and João Pedro Silvestre and Paulo Tabuada},
      year={2025},
      eprint={2412.02682},
      archivePrefix={arXiv},
      primaryClass={cs.AI},
      url={https://arxiv.org/abs/2412.02682}, 
}

@misc{goh2022simplicialattentionnetworks,
      title={Simplicial Attention Networks}, 
      author={Christopher Wei Jin Goh and Cristian Bodnar and Pietro Liò},
      year={2022},
      eprint={2204.09455},
      archivePrefix={arXiv},
      primaryClass={cs.LG},
      url={https://arxiv.org/abs/2204.09455}, 
}

@misc{veličković2018graphattentionnetworks,
      title={Graph Attention Networks}, 
      author={Petar Veličković and Guillem Cucurull and Arantxa Casanova and Adriana Romero and Pietro Liò and Yoshua Bengio},
      year={2018},
      eprint={1710.10903},
      archivePrefix={arXiv},
      primaryClass={stat.ML},
      url={https://arxiv.org/abs/1710.10903}, 
}

@misc{openai2024gpt4technicalreport,
      title={GPT-4 Technical Report}, 
      author={GPT-4},
      year={2024},
      eprint={2303.08774},
      archivePrefix={arXiv},
      primaryClass={cs.CL},
      url={https://arxiv.org/abs/2303.08774}, 
}

@misc{hajij2023topologicaldeeplearninggoing,
      title={Topological Deep Learning: Going Beyond Graph Data}, 
      author={Mustafa Hajij and Ghada Zamzmi and Theodore Papamarkou and Nina Miolane and Aldo Guzmán-Sáenz and Karthikeyan Natesan Ramamurthy and Tolga Birdal and Tamal K. Dey and Soham Mukherjee and Shreyas N. Samaga and Neal Livesay and Robin Walters and Paul Rosen and Michael T. Schaub},
      year={2023},
      eprint={2206.00606},
      archivePrefix={arXiv},
      primaryClass={cs.LG},
      url={https://arxiv.org/abs/2206.00606}, 
}

@misc{battaglia2018relationalinductivebiasesdeep,
      title={Relational inductive biases, deep learning, and graph networks}, 
      author={Peter W. Battaglia and Jessica B. Hamrick and Victor Bapst and Alvaro Sanchez-Gonzalez and Vinicius Zambaldi and Mateusz Malinowski and Andrea Tacchetti and David Raposo and Adam Santoro and Ryan Faulkner and Caglar Gulcehre and Francis Song and Andrew Ballard and Justin Gilmer and George Dahl and Ashish Vaswani and Kelsey Allen and Charles Nash and Victoria Langston and Chris Dyer and Nicolas Heess and Daan Wierstra and Pushmeet Kohli and Matt Botvinick and Oriol Vinyals and Yujia Li and Razvan Pascanu},
      year={2018},
      eprint={1806.01261},
      archivePrefix={arXiv},
      primaryClass={cs.LG},
      url={https://arxiv.org/abs/1806.01261}, 
}

@misc{ying2021transformersreallyperformbad,
      title={Do Transformers Really Perform Bad for Graph Representation?}, 
      author={Chengxuan Ying and Tianle Cai and Shengjie Luo and Shuxin Zheng and Guolin Ke and Di He and Yanming Shen and Tie-Yan Liu},
      year={2021},
      eprint={2106.05234},
      archivePrefix={arXiv},
      primaryClass={cs.LG},
      url={https://arxiv.org/abs/2106.05234}, 
}

@misc{bechlerspeicher2025positiongraphlearninglose,
      title={Position: Graph Learning Will Lose Relevance Due To Poor Benchmarks}, 
      author={Maya Bechler-Speicher and Ben Finkelshtein and Fabrizio Frasca and Luis Müller and Jan Tönshoff and Antoine Siraudin and Viktor Zaverkin and Michael M. Bronstein and Mathias Niepert and Bryan Perozzi and Mikhail Galkin and Christopher Morris},
      year={2025},
      eprint={2502.14546},
      archivePrefix={arXiv},
      primaryClass={cs.LG},
      url={https://arxiv.org/abs/2502.14546}, 
}

@misc{oquab2024dinov2learningrobustvisual,
      title={DINOv2: Learning Robust Visual Features without Supervision}, 
      author={Maxime Oquab and Timothée Darcet and Théo Moutakanni and Huy Vo and Marc Szafraniec and Vasil Khalidov and Pierre Fernandez and Daniel Haziza and Francisco Massa and Alaaeldin El-Nouby and Mahmoud Assran and Nicolas Ballas and Wojciech Galuba and Russell Howes and Po-Yao Huang and Shang-Wen Li and Ishan Misra and Michael Rabbat and Vasu Sharma and Gabriel Synnaeve and Hu Xu and Hervé Jegou and Julien Mairal and Patrick Labatut and Armand Joulin and Piotr Bojanowski},
      year={2024},
      eprint={2304.07193},
      archivePrefix={arXiv},
      primaryClass={cs.CV},
      url={https://arxiv.org/abs/2304.07193}, 
}

@misc{dosovitskiy2021imageworth16x16words,
      title={An Image is Worth 16x16 Words: Transformers for Image Recognition at Scale}, 
      author={Alexey Dosovitskiy and Lucas Beyer and Alexander Kolesnikov and Dirk Weissenborn and Xiaohua Zhai and Thomas Unterthiner and Mostafa Dehghani and Matthias Minderer and Georg Heigold and Sylvain Gelly and Jakob Uszkoreit and Neil Houlsby},
      year={2021},
      eprint={2010.11929},
      archivePrefix={arXiv},
      primaryClass={cs.CV},
      url={https://arxiv.org/abs/2010.11929}, 
}

@misc{radford2021learningtransferablevisualmodels,
      title={Learning Transferable Visual Models From Natural Language Supervision}, 
      author={Alec Radford and Jong Wook Kim and Chris Hallacy and Aditya Ramesh and Gabriel Goh and Sandhini Agarwal and Girish Sastry and Amanda Askell and Pamela Mishkin and Jack Clark and Gretchen Krueger and Ilya Sutskever},
      year={2021},
      eprint={2103.00020},
      archivePrefix={arXiv},
      primaryClass={cs.CV},
      url={https://arxiv.org/abs/2103.00020}, 
}

@misc{morris2021weisfeilerlemanneuralhigherorder,
      title={Weisfeiler and Leman Go Neural: Higher-order Graph Neural Networks}, 
      author={Christopher Morris and Martin Ritzert and Matthias Fey and William L. Hamilton and Jan Eric Lenssen and Gaurav Rattan and Martin Grohe},
      year={2021},
      eprint={1810.02244},
      archivePrefix={arXiv},
      primaryClass={cs.LG},
      url={https://arxiv.org/abs/1810.02244}, 
}

@misc{liang2024tensorattentiontrainingprovably,
      title={Tensor Attention Training: Provably Efficient Learning of Higher-order Transformers}, 
      author={Yingyu Liang and Zhenmei Shi and Zhao Song and Yufa Zhou},
      year={2024},
      eprint={2405.16411},
      archivePrefix={arXiv},
      primaryClass={cs.LG},
      url={https://arxiv.org/abs/2405.16411}, 
}

@misc{alman2023capturehigherordercorrelationsgeneralizing,
      title={How to Capture Higher-order Correlations? Generalizing Matrix Softmax Attention to Kronecker Computation}, 
      author={Josh Alman and Zhao Song},
      year={2023},
      eprint={2310.04064},
      archivePrefix={arXiv},
      primaryClass={cs.DS},
      url={https://arxiv.org/abs/2310.04064}, 
}

@misc{hussain2024tripletinteractionimprovesgraph,
      title={Triplet Interaction Improves Graph Transformers: Accurate Molecular Graph Learning with Triplet Graph Transformers}, 
      author={Md Shamim Hussain and Mohammed J. Zaki and Dharmashankar Subramanian},
      year={2024},
      eprint={2402.04538},
      archivePrefix={arXiv},
      primaryClass={cs.LG},
      url={https://arxiv.org/abs/2402.04538}, 
}

@misc{liu2022gem2generationmolecularproperty,
      title={GEM-2: Next Generation Molecular Property Prediction Network by Modeling Full-range Many-body Interactions}, 
      author={Lihang Liu and Donglong He and Xiaomin Fang and Shanzhuo Zhang and Fan Wang and Jingzhou He and Hua Wu},
      year={2022},
      eprint={2208.05863},
      archivePrefix={arXiv},
      primaryClass={cs.LG},
      url={https://arxiv.org/abs/2208.05863}, 
}

@article{Jumper2021,
  author = {John Jumper and Richard Evans and Alexander Pritzel and Tim Green and Michael Figurnov and Olaf Ronneberger and Kathryn Tunyasuvunakool and Russ Bates and Augustin \v{Z}ídek and Anna Potapenko and Alex Bridgland and Clemens Meyer and Simon A. A. Kohl and Andrew J. Ballard and Andrew Cowie and Bernardino Romera-Paredes and Stanislav Nikolov and Rishub Jain and Jonas Adler and Trevor Back and Stig Petersen and David Reiman and Ellen Clancy and Michal Zielinski and Martin Steinegger and Michalina Pacholska and Tamas Berghammer and Sebastian Bodenstein and David Silver and Oriol Vinyals and Andrew W. Senior and Koray Kavukcuoglu and Pushmeet Kohli and Demis Hassabis},
  title = {Highly accurate protein structure prediction with AlphaFold},
  journal = {Nature},
  year = {2021},
  url = {https://doi.org/10.1038/s41586-021-03819-2},
}

@misc{ballester2024attendingtopologicalspacescellular,
      title={Attending to Topological Spaces: The Cellular Transformer}, 
      author={Rubén Ballester and Pablo Hernández-García and Mathilde Papillon and Claudio Battiloro and Nina Miolane and Tolga Birdal and Carles Casacuberta and Sergio Escalera and Mustafa Hajij},
      year={2024},
      eprint={2405.14094},
      archivePrefix={arXiv},
      primaryClass={cs.LG},
      url={https://arxiv.org/abs/2405.14094}, 
}

@misc{zhou2024theoreticalexpressivepowerdesign,
      title={On the Theoretical Expressive Power and the Design Space of Higher-Order Graph Transformers}, 
      author={Cai Zhou and Rose Yu and Yusu Wang},
      year={2024},
      eprint={2404.03380},
      archivePrefix={arXiv},
      primaryClass={cs.LG},
      url={https://arxiv.org/abs/2404.03380}, 
}

@misc{castin2024smoothattention,
      title={How Smooth Is Attention?}, 
      author={Valérie Castin and Pierre Ablin and Gabriel Peyré},
      year={2024},
      eprint={2312.14820},
      archivePrefix={arXiv},
      primaryClass={cs.LG},
      url={https://arxiv.org/abs/2312.14820}, 
}

@misc{kim2021lipschitzconstantselfattention,
      title={The Lipschitz Constant of Self-Attention}, 
      author={Hyunjik Kim and George Papamakarios and Andriy Mnih},
      year={2021},
      eprint={2006.04710},
      archivePrefix={arXiv},
      primaryClass={stat.ML},
      url={https://arxiv.org/abs/2006.04710}, 
}

@misc{federer2014geometric,
  title     = {Geometric Measure Theory},
  author    = {Federer, Herbert},
  publisher = {Springer Berlin Heidelberg},
  year      = {2014},
  isbn      = {9783642620102},
  series    = {Classics in Mathematics},
  url       = {https://books.google.fr/books?id=jld-BgAAQBAJ}
}

@misc{zhang2019rootmeansquarelayer,
      title={Root Mean Square Layer Normalization}, 
      author={Biao Zhang and Rico Sennrich},
      year={2019},
      eprint={1910.07467},
      archivePrefix={arXiv},
      primaryClass={cs.LG},
      url={https://arxiv.org/abs/1910.07467}, 
}

@misc{ba2016layernormalization,
      title={Layer Normalization}, 
      author={Jimmy Lei Ba and Jamie Ryan Kiros and Geoffrey E. Hinton},
      year={2016},
      eprint={1607.06450},
      archivePrefix={arXiv},
      primaryClass={stat.ML},
      url={https://arxiv.org/abs/1607.06450}, 
}
